\documentclass[letterpaper]{article} 
\usepackage[]{aaai2026}  
\usepackage{times}  
\usepackage{helvet}  
\usepackage{courier}  
\usepackage[hyphens]{url}  
\usepackage{graphicx} 
\usepackage{natbib}  
\usepackage{caption} 
\usepackage{algorithm}
\usepackage{algorithmic}
\usepackage{enumitem}
\usepackage{amsmath} 
\usepackage{amssymb}  
\usepackage{mathrsfs} 
\usepackage{booktabs}
\usepackage{multirow}
\nocopyright  
\usepackage{newfloat}
\usepackage{listings}
\DeclareCaptionStyle{ruled}{labelfont=normalfont,labelsep=colon,strut=off} 
\floatstyle{ruled}
\newfloat{listing}{tb}{lst}{}
\floatname{listing}{Listing}
\title{CM-Diff: A Single Generative Network for Bidirectional Cross-Modality Translation Diffusion Model Between Infrared and Visible Images}
\author {
    Bin Hu\textsuperscript{\rm 1} \quad
    Chenqiang Gao\textsuperscript{\rm 1,2*} \quad
    Shurui Liu\textsuperscript{\rm 2} \quad
    Fang Chen\textsuperscript{\rm 3} \\
    Junjie Guo\textsuperscript{\rm 1} \quad
    Fangcen Liu\textsuperscript{\rm 1} \quad
    Junwei Han\textsuperscript{\rm 1,4}
}

\affiliations {
    \textsuperscript{\rm 1}Chongqing University of Posts and Telecommunications \quad
    \textsuperscript{\rm 2}Sun Yat-sen University \\
    \textsuperscript{\rm 3}University of California, Merced \quad
    \textsuperscript{\rm 4}Northwestern Polytechnical University \\
    {\tt\small \{haydengauthiero,liufc67,gjj893866738,liushr6688\}@gmail.com, gaochq6@mail.sysu.edu.cn,} \\
    {\tt\small fchen20@ucmerced.edu, jhan@nwpu.edu.cn} 
}

\usepackage{bibentry}

\begin{document}
\maketitle

\begin{abstract}
Image translation is one of the crucial approaches for mitigating information deficiencies in the infrared and visible modalities, while also facilitating the enhancement of modality-specific datasets.
However, existing methods for infrared and visible image translation either achieve unidirectional modality translation or rely on cycle consistency for bidirectional modality translation, which may result in suboptimal performance.
In this work, we present the bidirectional cross-modality translation diffusion model (CM-Diff) for simultaneously modeling data distributions in both the infrared and visible modalities.
We address this challenge by combining translation direction labels for guidance during training with cross-modality feature control.
Specifically, we view the establishment of the mapping relationship between the two modalities as the process of learning data distributions and understanding modality differences, achieved through a novel Bidirectional Diffusion Training (BDT).
Additionally, we propose a Statistical Constraint Inference (SCI) to ensure the generated image closely adheres to the data distribution of the target modality.
Experimental results demonstrate the superiority of our CM-Diff over state-of-the-art methods, highlighting its potential for generating dual-modality datasets.
\end{abstract}

\section{Introduction}
\begin{figure}[!ht]
    \centering
    \includegraphics[scale=0.310, trim=31mm 157mm 28mm 54mm, clip, width=0.48\textwidth]{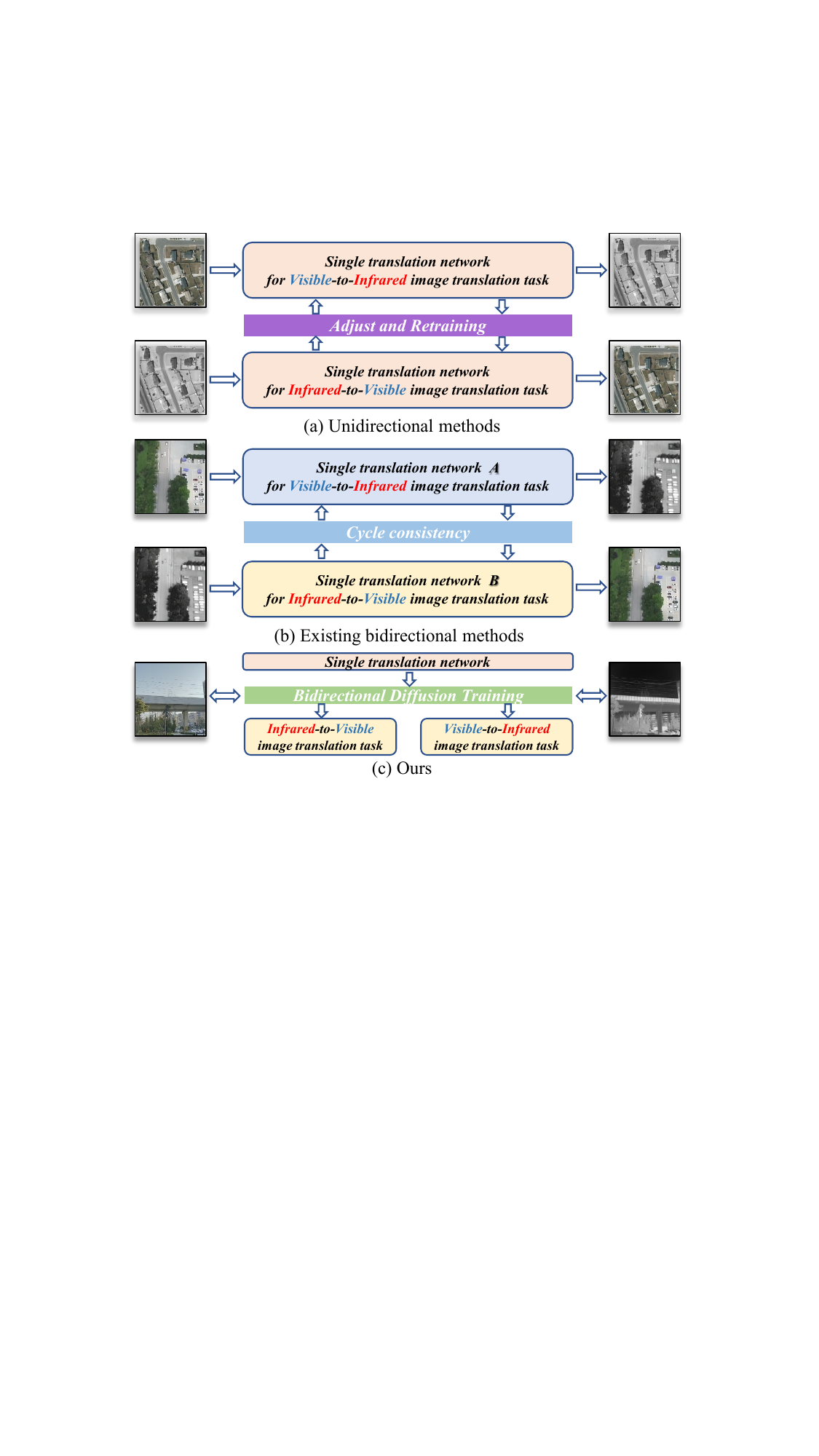}
\caption{Comparison of infrared-visible image translation methods: (a) Unidirectional methods require network modifications and retraining to enable translation in the opposite direction; (b) Bidirectional methods employ two independent generative networks and utilize cycle consistency for simultaneous training; (c) CM-Diff introduces Bidirectional Diffusion Training (BDT), eliminating the need for separate networks and reducing the computational overhead and retraining time associated with unidirectional methods.
}
\label{figure:intro}
\end{figure}
Infrared (IR) and visible (VIS) imaging are two common modalities in visual perception within the computer vision community. 
Infrared images are captured on the basis of the thermal radiation emitted by objects, which makes them invariant to illumination conditions and robust against environmental disturbances such as low light and smoke \cite{infmae}. 
Nevertheless, they lack color, texture, and detailed information, which limits their ability to analyze fine-grained objects. 
On the other hand, visible images are formed by the reflection of light from objects, providing rich color, texture, and detailed information, enabling precise object representation and fine-grained analysis. 
This disparity in the characteristics of infrared and visible imaging creates a complementary relationship between the two modalities, making their integration or alternation crucial for robust real-world applications. 
Given the complementary strengths and limitations of visible and infrared imaging, dual-modality methods often outperform single-modality methods, particularly in tasks like object detection \cite{damsdet, fu2023lraf, li2023stabilizing} and semantic segmentation \cite{liu2023multi, ji2023multispectral, deng2021feanet}.
However, these methods usually rely on a large, well-aligned dual-modality dataset, while in practice, acquiring such registered dual-modality datasets is challenging and costly. 
Thus, a natural idea is to leverage image translation techniques to generate dual-modality datasets.

Recent infrared-visible image translation methods have achieved promising results using Generative Adversarial Networks (GANs) \cite{liu2018ir2vi, wang2024mappingformer, han2024dr, Tirdet, polarimetric, TVA-GAN, DR-AVIT} and Denoising Diffusion Probabilistic Models (DDPMs) \cite{LG-Diff, Unit-ddpm, T2v-ddpm, difftv, oliveira2024stable}, which  can be broadly classified into two categories: bidirectional image translation methods and unidirectional image translation methods. 
Bidirectional methods \cite{CycleGAN, UNIT, DCLGAN, Tirdet, Unit-ddpm} typically employ two independent generative networks to establish mappings between the infrared and visible modalities, often relying on a cycle consistency method, as shown in Fig. \ref{figure:intro} (b).
Although the cycle consistency based methods have proven to be effective in ensuring modality correspondence, it not only increases the complexity of the network architecture but also frequently leads to perceptual blurriness and loss of fine details in the generated outputs \cite{CycleGAN, maqsoodcycle}. 
Unidirectional methods \cite{Tv-gan, polarimetric, Pix2Pix, TVA-GAN, ma2024visible, Ic-gan, DR-AVIT} leverage a single generative network to model the mapping in one direction, either infrared-to-visible image translation or vice versa. 
Consequently, they require separate training processes for each translation direction to achieve bidirectional modality translation, as shown in Fig. \ref{figure:intro} (a).

In this paper, we propose a unified framework called CM-Diff.
The core idea is to enable joint learning of infrared and visible image data using a Bidirectional Diffusion Training (BDT), as shown in Fig. \ref{figure:intro} (c).
This method incorporates implicit prompts through input channel positions, combined with explicit directional label embedding. 
This method is further refined by a multi-level Cross-modality Feature Control (CFC), which enhances semantic control over the translation process. Consequently, it eliminates reliance on the cycle-consistency method while enabling more precise mapping between the two data modalities.
Furthermore, we observe that after implementing bidirectional conversion, the issue of color distortion becomes critically severe when directly applying the original ancestral sampling method \cite{AncestralSampling, surveydiffusion} for infrared-visible image translation. 
To address this, we propose a Statistical Constraint Inference (SCI), which leverages statistical mean constraints tailored to different datasets and scenarios.
By incorporating these statistical properties, we mitigate abnormalities caused by the inherent randomness in the reverse Markov chain of DDPM, ensuring that the generated images maintain high fidelity and align closely with real-world distributions.
The main contributions of this paper can be summarized as follows:
\begin{enumerate}[label=\tiny\textbullet]
  \item \textbf{A unified CM-Diff framework for bidirectional IR–VIS translation.} We propose CM-Diff with a Bidirectional Diffusion Training (BDT) that encodes direction labels and source-domain features into a single U-Net, enabling IR→VIS and VIS→IR mappings without dual networks or additional consistency losses.
  \item \textbf{Statistical Constraint Inference (SCI) for distributional alignment.} By integrating domain-specific statistics (e.g., thermal variance) into the reverse diffusion process, SCI corrects sampling drift and aligns the generated image distribution with real data, effectively reducing anomalies and color distortion.
  \item \textbf{Extensive experimental validation.} Comprehensive experiments on standard benchmarks demonstrate that our method outperforms existing approaches in both quantitative metrics and qualitative visual consistency, addressing modality absence and misalignment issues.
\end{enumerate}

\section{Related Work}
Cross-modality translation between infrared and visible images is a challenging task due to the significant modality gap in texture, color, and semantics. Existing approaches can be broadly categorized into GAN-based, DDPM-based, and more recent bidirectional generative modeling techniques.

\subsection{GAN-based Cross-Modality Image Translation}
Generative Adversarial Networks have been widely used for IR-VIS translation tasks, often leveraging pixel-level supervision or perceptual loss to bridge the modality gap. For IR$\rightarrow$VIS, TVA-GAN \cite{TVA-GAN} incorporates attention and inception-based features to improve local structural detail, while I2V-GAN \cite{I2v-gan} adds cyclic and perceptual constraints to achieve temporal coherence in video translation. ROMA \cite{ROMA} and EADS \cite{EADS} enhance semantic alignment by integrating edge or similarity priors. In contrast, VIS$\rightarrow$IR translation focuses on thermal realism. For example, VTF-GAN \cite{When} introduces Fourier losses to recover frequency-aware details, and IC-GAN \cite{Ic-gan} employs color-space mapping to reduce appearance distortion in remote sensing.

While GANs produce sharp outputs, they are prone to mode collapse and training instability. More importantly, most methods are unidirectional, requiring separate training for each direction. Bidirectional translation using GANs typically relies on two models with cycle consistency (e.g., CycleGAN variants), which adds training overhead and may degrade translation quality.

\subsection{DDPM-based Cross-Modality Image Translation}
Diffusion models offer an alternative that avoids GAN-specific instabilities by gradually denoising from noise via a Markov chain. T2V-DDPM \cite{T2v-ddpm} pioneers their use in IR-to-VIS face translation under low-light conditions, while UNIT-DDPM \cite{Unit-ddpm} jointly models both modalities through dual-domain Markov chains. However, these methods remain unidirectional and require retraining to switch translation direction.

Some diffusion-based editing methods support partial reversibility. SDEdit \cite{sdedit}, for instance, enables image synthesis through partial denoising but assumes conditional masks and is primarily used in unpaired settings. EGSDE \cite{egsde} introduces energy-guided sampling, yet it remains unpaired and task-specific. These techniques are not directly comparable due to assumptions not holding in our paired IR-VIS domain.

\subsection{Bidirectional Diffusion and Bridge Models}
To eliminate the need for dual generators or cycle loss, several works propose bidirectional learning within a single model. Bidirectional Consistency Models (BCM) \cite{bcm} introduce a consistency loss to regularize one diffusion model for both directions, but they operate in unpaired domains. In contrast, our setting assumes paired supervision, which allows stronger data-driven training signals.

Similarly, Cycle-Free CycleGAN \cite{cyclefree-cyclegan} avoids cycle loss by using an invertible generator in medical imaging, which still differs architecturally and assumes invertibility—an assumption hard to satisfy in IR-VIS tasks due to non-invertible thermal content. Schrödinger Bridge methods such as I²SB \cite{i2sb} and CBD \cite{cbd} formulate bidirectional generation as stochastic transport, relying on bridging distributions. These methods are theoretically elegant but often require iterative sampling or multi-step alignment and are more suited to unsupervised or unpaired tasks.

Unlike these approaches, our method is the first to enable bidirectional IR$\leftrightarrow$VIS generation using a single shared diffusion model. The direction of translation is encoded via the timestep embedding, without requiring auxiliary supervision or additional inference passes.

\section{Method}
\begin{figure*}[ht]
    \centering
    \includegraphics[scale=0.1, trim=6mm 218mm 6.95mm 51.5mm, clip, width=1\textwidth]{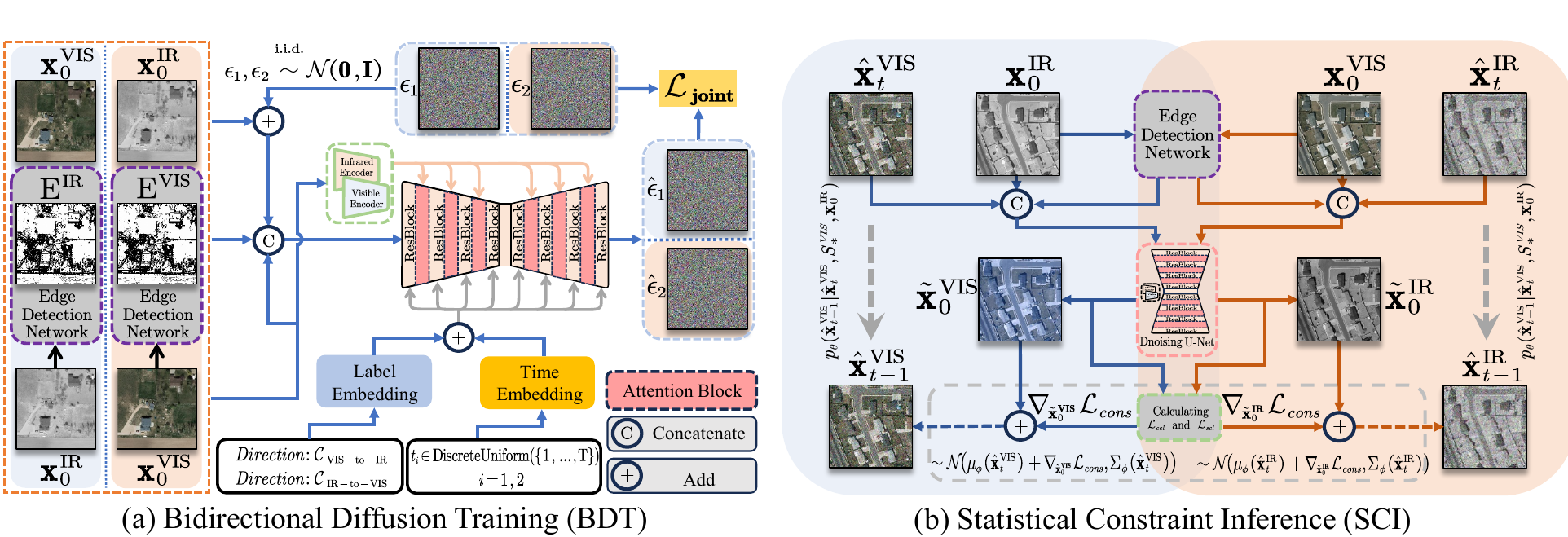}
    \caption{
   {Overview of CM-Diff.} (a) {Bidirectional Diffusion Training (BDT) :} Learning domain-specific data distributions and establishing robust bidirectional mappings. (b) {Statistical Constraint Inference (SCI) :} Ensuring consistency between the statistical distributions of translated images and those in the training data.
}
    \label{figure:model}
\end{figure*} 
Given a set of paired infrared images \(\mathcal{X}^{\textsc{ir}} = \{\mathbf{x}^{\textsc{ir}}_{1}, \cdots, \mathbf{x}^{\textsc{ir}}_{N} \mid \mathbf{x} \in \mathbb{R}^{H\times W \times C_{\textsc{ir}}} \}\) and visible images \(\mathcal{X}^{\textsc{vis}} = \{\mathbf{x}^{\textsc{vis}}_{1}, \cdots, \mathbf{x}^{\textsc{vis}}_{N} \mid \mathbf{x} \in \mathbb{R}^{H\times W \times C_{\textsc{vis}}} \}\), our goal is to train a model  $\mathcal{Z}_{N}$ capable of translating infrared images \(\mathcal{X}^{\textsc{ir}}\) into their corresponding visible counterparts \(\hat{\mathcal{X}}^{\textsc{vis}}\), while also being able to perform the reverse translation from visible images \(\mathcal{X}^{\textsc{vis}}\) to infrared images \(\hat{\mathcal{X}}^{\textsc{ir}}\).
To accomplish this task, we propose a bidirectional end-to-end framework, CM-Diff.
\subsection{Bidirectional Diffusion Training (BDT)}
Paired infrared (IR)  and visible (VIS) images contain similar semantic information, but their data distributions differ significantly due to the distinct imaging characteristics of each modality. 
To simultaneously learn these two data distributions and establish a bidirectional mapping relationship, we propose a Bidirectional Diffusion Training (BDT). 
Our strategy uses a U-Net architecture network, as shown in Fig. \ref{figure:model} (a), to simultaneously learn and distinguish the visible modality data $\mathbf{x}_{0}^{\textsc{vis}} \in \mathcal{X}^{\textsc{vis}}$ and infrared modality data $\mathbf{x}_{0}^{\textsc{ir}} \in \mathcal{X}^{\textsc{ir}}$, and institutes an effective bidirectional mapping to ensure that the content of the source modality image corresponds effectively to the translated target modality image, thereby achieving a stable training process.
\paragraph{(1) Translation Direction Guidance (TDG).} 
 We first employ an implicit input channel position encoding to determine whether the data distribution being learned is from the infrared modality or the visible modality. Specifically, for the infrared-to-visible translation part, the input consists of three components: the noisy data $q\left( \mathbf{x}^{\textsc{vis}}_{t} \mid \mathbf{x}^{\textsc{vis}}_{0} \right) \in \mathbb{R}^{ H \times W \times C_{\textsc{vis}}}$ in the visible modality at time step $t$, the infrared modality image data $\mathbf{x}^{\textsc{ir}}_{0} \in\mathbb{R}^{ H \times W \times C_{\textsc{ir}}}$, and the edge feature maps $E^{\textsc{ir}} = \text{DexiNed}\left( \mathbf{x}^{\textsc{ir}} \right) \in \mathbb{R}^{ H \times W \times E}$ extracted from the infrared image using the edge detection network DexiNed \cite{Dense}. 
For the consistency of subsequent operations, we duplicate the single-channel infrared image across three channels to match the visible format, ensuring $C_{\textsc{vis}} = C_{\textsc{ir}} = 3$. Using edge feature maps extracted from infrared and visible images as part of the input prompt can ensure the consistency of the target image \cite{EADS}.
To construct the network input, we concatenate these data along the channel dimension, resulting in the following input representation:
\begin{equation}
\mathcal{Z}_{\text{\textsc{ir}-to-\textsc{vis}}}^{t} = q\left( \mathbf{x}_{t}^{\textsc{vis}} \mid \mathbf{x}^{\textsc{vis}}_{0} \right) \oplus \mathbf{x}^{\textsc{ir}}_{0} \oplus E^{\textsc{ir}},
\end{equation}
where $\mathcal{Z}_{\text{\textsc{ir}-to-\textsc{vis}}}^{t} \in \mathbb{R}^{ H \times W \times \left(C_{\textsc{vis}} + C_{\textsc{ir}} + E \right)}$, $\oplus$ denotes the concatenation operation, with the target modality data, source modality data, and edge features corresponding to channel index ranges $\{1, 2, \cdots, C_{\textsc{vis}}\}$, $\{C_{\textsc{vis}}+1, C_{\textsc{vis}}+2, \cdots, C_{\textsc{ir}} + C_{\textsc{vis}}\}$, and $\{C_{\textsc{ir}} + C_{\textsc{vis}}+1, C_{\textsc{ir}} + C_{\textsc{vis}}+2, \cdots, C_{\textsc{ir}} + C_{\textsc{vis}}+E\}$, respectively. This implicit channel encoding effectively distinguishes the data from different modalities without requiring explicit direction labels indicating the data source.
Additionally, we introduce label embedding layers $\tau_{\phi}$ to process the translation direction labels, which represent: $\mathcal{C}^{\textsc{ir}}$ corresponding to the image translation direction from visible to infrared, learning the data distribution of the infrared modality.
\paragraph{(2)  Cross-modality Feature Control (CFC).} 
 To achieve effective target-modality image content control based on the source-modality image, in addition to the direct input of edge feature maps, we also train two independent U-Net encoders, one serving as the infrared modality encoder $\mathcal{Z}_{E}^{\textsc{ir}}$ and the other as the visible modality encoder $\mathcal{Z}_{E}^{\textsc{vis}}$, to extract modality features $F_{d}$ from the infrared or visible images. These modality features are then fed into the cross-attention layer of $\mathcal{Z}_{N}$ to enhance the control of the target modality image content. The query $Q_{n}^{i}$, key $K_{n}^{i}$, and value $V_{n}^{i}$ are computed as follows:
\begin{equation}
    Q_{n}^{i} = Conv_{q}^{i}\left( F_{g_{i}}^{n} \right), \quad K_{n}^{i}, V_{n}^{i} = Conv_{kv}^{i}\left( F_{d_{i}}^{n} \right),
\end{equation}
where $F_{g_{i}}^{n}$ is the input vector obtained by flattening the noise feature map of the $i$-th cross-attention part in the $n$-th layer of network $\mathcal{Z}_{N}$, and $F_{d_{i}}^{n}$ is the feature map extracted from the $i$-th part in the $n$-th layer of the modality encoder.
Through the cross-modality attention mechanism, we compute the attention weights $A^{i}$ and obtain the updated feature output $F_{g_{i+1}}^{n}$:
\begin{equation}
    F_{g_{i+1}}^{n} = F_{g_{i}}^{n} + {softmax}\left( \frac{Q_{n}^{i} \cdot \left(K_{n}^{i}\right)^{T}}{\sqrt{d}} \right) \cdot V_{n}^{i}.
\end{equation}
\paragraph{(3) Modality Joint Learning Loss Function.} 
Once the model effectively learns to capture the characteristics of both data distributions, we compute the noise prediction loss in each translation direction.
Specifically, when the infrared image provides guidance, the mean squared error (MSE) between the predicted noise in the noisy visible image and the true noise is calculated, yielding $  \mathcal{L}_{\text{\textsc{ir}-to-\textsc{vis}}} $. Conversely,when the visible image provides guidance, the corresponding MSE is computed for the infrared image, resulting in $\mathcal{L}_{\text{\textsc{vis}-to-\textsc{ir}}}$ . The loss functions are defined as:
\begin{equation}
\begin{aligned}
    &\mathcal{L}_{\text{\textsc{ir}-to-\textsc{vis}}}\left( \phi \right) = \mathbb{E}_{t_{1}, \mathbf{x}_{0}^{\textsc{ir}}, \epsilon_{1}} \left[ \left\| \epsilon_{1} - \epsilon_{\phi} \left( \mathcal{Z}_{\text{\textsc{ir}-to-\textsc{vis}}}^{t_{1}}, J^{\textsc{vis}}, t_{1} \right) \right\|_{2}^{2} \right] \\
    &\mathcal{L}_{\text{\textsc{vis}-to-\textsc{ir}}}\left( \phi \right) = \mathbb{E}_{t_{2}, \mathbf{x}_{0}^{\textsc{vis}}, \epsilon_{2}} \left[ \left\| \epsilon_{2} - \epsilon_{\phi} \left( \mathcal{Z}_{\text{\textsc{vis}-to-\textsc{ir}}}^{t_{2}}, J^{\textsc{ir}}, t_{2} \right) \right\|_{2}^{2} \right],
\end{aligned}
\end{equation}
where $t_1, t_2 \overset{\text{i.i.d.}}{\sim} \mathrm{DiscreteUniform}(\{1, \cdots, T\})$, $\epsilon_1, \epsilon_2 \overset{\text{i.i.d.}}{\sim} \mathcal{N}\left( \mathbf{0}, \mathbf{I} \right)$,  $J = \tau_{\phi} \left( \mathcal{C} \right)$ . The modality joint learning loss $\mathcal{L}_{{joint}}$ is as follows:
\begin{equation}
    \mathcal{L}_{{joint}} \left( \phi \right) = \lambda_{\text{\textsc{ir}-to-\textsc{vis}}} \cdot \mathcal{L}_{\text{\textsc{ir}-to-\textsc{vis}}} + \lambda_{\text{\textsc{vis}-to-\textsc{ir}}} \cdot \mathcal{L}_{\text{\textsc{vis}-to-\textsc{ir}}}.
\end{equation}
\begin{algorithm}[th]
\renewcommand{\algorithmicrequire}{\textbf{Input:}}
\renewcommand{\algorithmicensure}{\textbf{Output:}}
	\caption{CM-Diff Training}
    \label{Algorithm1}
    \begin{algorithmic}[1]
        \REQUIRE  visible (VIS) image $\mathbf{x}^{\textsc{vis}}$,  infrared (IR) image $\mathbf{x}^{\textsc{ir}}$, infrared-to-visible direction label $\mathcal{C}^{\textsc{ir}}$, visible-to-infrared direction label $\mathcal{C}^{\textsc{vis}}$.
        \REPEAT
            \STATE  $\mathbf{x}^{\textsc{vis}} \in \mathcal{X}^{\textsc{vis}},\mathbf{x}^{\textsc{ir}} \in \mathcal{X}^{\textsc{ir}}$
            \STATE  $t_1, t_2 \overset{\text{i.i.d.}}{\sim} \mathrm{DiscreteUniform}(\{1, \cdots, T\})$
            \STATE  $\epsilon _{1}, \epsilon _{2} \overset{\text{i.i.d.}}{\sim} \mathcal N\left (\mathbf{0},\mathbf{I}  \right ) $
            \STATE  $\mathbf{x}^{\textsc{vis}}_{t_{1}}\gets \sqrt[]{{\bar{\alpha}}_{t_{1}}} \cdot \mathbf{x}_{0}^{\textsc{vis}}+\sqrt{1-\bar{\alpha}_{t_{1}}} \cdot \epsilon_{1}$ \\
                    $\mathbf{x}^{\textsc{ir}}_{t_{2}}\gets \sqrt[]{{\bar{\alpha}}_{t_{2}}} \cdot \mathbf{x}_{0}^{\textsc{ir}}+\sqrt{1-\bar{\alpha}_{t_{2}}} \cdot \epsilon _{2}$
            \STATE  $J^{\textsc{vis}} \gets \tau_{\phi }(\text{$\mathcal{C}^{\textsc{vis}}$}), J^{\textsc{ir}} \gets \tau_{\phi }(\text{$\mathcal{C}^{\textsc{ir}}$})$ \\
            \STATE  $E^{\textsc{vis}} \gets \text{DexiNed}\left(\mathbf{x}_{0}^{\textsc{vis}}\right), E^{\textsc{ir}} \gets 
            \text{DexiNed}\left(\mathbf{x}_{0}^{\textsc{ir}}\right)$ \\
            \STATE  Gradient descent step on: \\ $\nabla _{\phi }\left [ \lambda_{\text{\textsc{ir}-to-\textsc{vis}}} \cdot \left \| \epsilon_{1} -\epsilon _{\phi  }\left ( \mathcal{Z}_{\text{\textsc{ir}-to-\textsc{vis}}}^{t_{1}},J^{\textsc{vis}},t_{1}  \right ) \right \|_{2}^{2} \right.\newline \left.
            \quad \hspace{0.05cm} + \lambda_{\text{\textsc{vis}-to-\textsc{ir}}} \cdot \left \| \epsilon_{2} -\epsilon _{\phi  }\left ( \mathcal{Z}_{\text{\textsc{vis}-to-\textsc{ir}}}^{t_{2}},J^{\textsc{ir}},t_{2}  \right ) \right \|_{2}^{2}  \right]$
        \UNTIL {Converged}
    \end{algorithmic}
\end{algorithm}
\subsection{Statistical Constraint Inference (SCI)}
Our statistical constraint inference (SCI) module aims to guide the IR$\rightarrow$VIS generation toward semantically and thermally consistent outputs.  
Related ideas include Diffusion Posterior Sampling (DPS) \cite{dps}, which constrains diffusion inference via approximate posteriors. However, SCI differs by using domain-specific statistics (e.g., thermal variance) instead of learned masks or latent priors, making it more suitable for infrared–visible translation where semantic cues are inherently weak.

In order to enhance bidirectional diffusion inference with cross-modality statistical awareness, we propose Statistical Constraint Inference (SCI), which constrains the denoising trajectory using prior statistical properties derived from paired training data. The method consists of three components:  
(1) baseline inference,  
(2) constrained denoising using statistical priors, and  
(3) loss-guided gradient-based sampling.

\paragraph{(1) Baseline Inference Formulation.} 
To adapt the denoising diffusion process to our Bidirectional Diffusion Training (BDT) framework, we reformulate the reverse transitions in each direction using conditional generative functions. Specifically, the denoising step is defined as:
\begin{equation}\label{15}
\begin{aligned}
    &\mathbf{x}_{t-1}^{\textsc{vis}} = \mu_{\phi} \left( \mathcal{Z}_{\text{\textsc{ir}-to-\textsc{vis}}}^{t_{1}}, J^{\textsc{vis}}, t_{1} \right) 
    + \Sigma_{\phi} \left( \mathcal{Z}_{\text{\textsc{ir}-to-\textsc{vis}}}^{t_{1}}, J^{\textsc{vis}}, t_{1} \right) \epsilon_{1},\\
    &\mathbf{x}_{t-1}^{\textsc{ir}}  = \mu_{\phi} \left( \mathcal{Z}_{\text{\textsc{vis}-to-\textsc{ir}}}^{t_{2}}, J^{\textsc{ir}}, t_{2} \right) 
    + \Sigma_{\phi} \left( \mathcal{Z}_{\text{\textsc{vis}-to-\textsc{ir}}}^{t_{2}}, J^{\textsc{ir}}, t_{2} \right) \epsilon_{2}.
\end{aligned}
\end{equation}
where $\mathcal{Z}_{\cdot}^{t}$ denotes the latent embedding extracted from the conditional encoder at time step $t$ along each direction, and $J^{\cdot}$ represents modality-specific guidance such as edge or semantic priors. The functions $\mu_{\phi}(\cdot)$ and $\Sigma_{\phi}(\cdot)$ produce the mean and variance of the denoising distribution at step $t$.

This formulation allows both visible and infrared modalities to be reconstructed from Gaussian noise by iteratively incorporating the complementary information provided by the opposite modality. Unlike unidirectional DDPM, our bidirectional structure ensures mutual enhancement and robust translation under strong structural constraints.

\paragraph{(2) Statistical-Constrained Reverse Sampling.}
To enhance semantic fidelity and statistical realism in generated images, we incorporate domain-specific statistical priors into the reverse denoising process by modifying the standard unconditional transition distribution $p_{\theta}(\mathbf{x}_{t-1} \mid \mathbf{x}_{t})$ into a constrained form $p_{\theta}(\mathbf{x}_{t-1} \mid \mathbf{x}_{t}, w)$, where $w$ encodes statistical properties of the target domain, such as edge distributions or pixel-level histograms.

Following the framework of classifier-free guidance \cite{GuidedDiffusion}, the conditional distribution can be approximated as:
\begin{equation}
\begin{aligned}
&\log p_{\theta }(\mathbf{x}_{t-1} \mid \mathbf{x}_{t}, w) =\log(p_{\theta }(\mathbf{x}_{t-1} \mid \mathbf{x}_{t})p(w \mid \mathbf{x}_{t}))+{z}_1 \\
&\approx \log p\left ( h \right ) + {z}_2, \quad h\sim \mathcal{N}\left ( \mu_{\theta} + g\Sigma_{\theta}, \Sigma_{\theta} \right ),
\end{aligned}
\end{equation}
where $g=\nabla _{\mathbf{x}_{t}}\log p\left ( w \mid \mathbf{x}_{t} \right ) $, ${z}_{1}$ and ${z}_{2}$ are constants.

However, directly estimating $p(w \mid \mathbf{x}_t)$ is unreliable because $\mathbf{x}_t$ is corrupted by time-dependent noise. Therefore, inspired by \cite{DDPM}, we introduce $\tilde{\mathbf{x}}_{0}$—an intermediate prediction of the clean image:
\begin{equation}\label{7}
    \tilde{\mathbf{x}}_{0}=\frac{1}{\sqrt[]{\bar{\alpha }_{t}} } \cdot \left ( \mathbf{x}_{t}-\sqrt{1-\bar{\alpha }_{t}} \cdot \epsilon _{\theta }(\mathbf{x}_{t}, t) \right ) .
\end{equation}
This prediction captures more reliable semantic content than $\mathbf{x}_t$, and is used as a proxy for evaluating $p(w \mid \tilde{\mathbf{x}}_0)$.

To construct a practical approximation of $p(w \mid \tilde{\mathbf{x}}_0)$, we define:
\begin{equation}
p(w \mid \tilde{\mathbf{x}}_0) = \frac{1}{\mathcal{M}} \exp \left\{ - \left[ \lambda_{\mathcal{Q}} \cdot \mathcal{Q}(\tilde{\mathbf{x}}_0, w) + \lambda_{\mathcal{K}} \cdot \mathcal{K}(\tilde{\mathbf{x}}_0) \right] \right\},
\end{equation}
where $\mathcal{Q}(\cdot)$ measures the distance between the statistics of $\tilde{\mathbf{x}}_0$ and the target domain (e.g., intensity histograms or edge maps), and $\mathcal{K}(\cdot)$ is a regularization term that penalizes visual artifacts such as blur or noise. The scale parameters $\lambda_{\mathcal{Q}}$ and $\lambda_{\mathcal{K}}$ control the relative influence of each constraint.

Accordingly, the gradient $g$ becomes:
\begin{equation}
\nabla_{\tilde{\mathbf{x}}_0} \log p(w \mid \tilde{\mathbf{x}}_0) = - \lambda_{\mathcal{Q}} \cdot \nabla_{\tilde{\mathbf{x}}_0} \mathcal{Q}(\tilde{\mathbf{x}}_0, w) - \lambda_{\mathcal{K}} \cdot \nabla_{\tilde{\mathbf{x}}_0} \mathcal{K}(\tilde{\mathbf{x}}_0).
\end{equation}

By injecting the guidance term $\Sigma_{\theta} \cdot \nabla_{\tilde{\mathbf{x}}_0} \log p(w \mid \tilde{\mathbf{x}}_0)$ into the reverse process, we steer the denoising trajectory toward outputs that respect both structural fidelity and global statistical properties, while maintaining the stochastic integrity of the original diffusion model.

Finally, the updated prediction of $\mathbf{x}_{t-1}$ is derived from the posterior distribution:
\begin{equation}\label{9}
\begin{aligned}
&q\left(\mathbf{x}_{t-1} \mid \mathbf{x}_{t} , \tilde{\mathbf{x}}_{0}\right) = \mathcal{N} \left(\mathbf{x}_{t-1}; \mu_{q}\left( \mathbf{x}_{t} , \tilde{\mathbf{x}}_{0}\right), \Sigma_{q}\left(t\right) \right), \\
&\mu_{q}\left( \mathbf{x}_{t} , \tilde{\mathbf{x}}_{0}\right)=\frac{\left(1-\alpha_{t}\right) \cdot \left(1-\bar{\alpha}_{t-1}\right) \cdot \mathbf{x}_{t}+\sqrt{\bar{\alpha}_{t-1}} \cdot \beta_{t} \cdot \tilde{\mathbf{x}}_{0}}{1-\bar{\alpha}_{t}},\\
&\Sigma_{q}\left(t\right) =\frac{\left(1-\alpha_{t}\right) \cdot \left(1-\bar{\alpha}_{t-1}\right)}{1-\bar{\alpha}_{t}} \cdot \mathbf{I}.
\end{aligned}
\end{equation}
The overall process of the SCI is presented in Fig. \ref{figure:model} (b).
\begin{table*}[ht]
\caption[]{Quantitative comparison of the CM-Diff with state-of-the-art methods. The best performance for each metric is highlighted in bold, and the second-best scores are marked with underline fonts for reference.} \label{table1}
\centering
\resizebox{\textwidth}{!}{ 
\begin{tabular}{l|cccc|cccc|cccc|cccc|cccc|cccc}
\toprule
\multirow{2}{*}{\textbf{Method}} & \multicolumn{4}{c|}{AVIID( \textbf{infrared $\rightarrow$ visible})} & \multicolumn{4}{c|}{VEDAI(\textbf{infrared $\rightarrow$ visible})} & \multicolumn{4}{c|}{$\text{M}^{3}\text{FD}$(\textbf{infrared $\rightarrow$ visible})} & \multicolumn{4}{c|}{AVIID( \textbf{visible$\rightarrow$ infrared })}& \multicolumn{4}{c|}{VEDAI( \textbf{visible$\rightarrow$ infrared })}& \multicolumn{4}{c}{$\text{M}^{3}\text{FD}$( \textbf{visible$\rightarrow$ infrared })}\\
~ & PSNR$\uparrow$ & SSIM$\uparrow$ & LPIPS$\downarrow$ & FID$\downarrow$   & PSNR$\uparrow$ & SSIM$\uparrow$ & LPIPS$\downarrow$ &FID$\downarrow$ 
 & PSNR$\uparrow$ & SSIM$\uparrow$ & LPIPS$\downarrow$ &FID$\downarrow$ 
 & PSNR$\uparrow$ & SSIM$\uparrow$ & LPIPS$\downarrow$ & FID$\downarrow$   & PSNR$\uparrow$ & SSIM$\uparrow$ & LPIPS$\downarrow$ & FID$\downarrow$ 
& PSNR$\uparrow$ & SSIM$\uparrow$ & LPIPS$\downarrow$ &FID$\downarrow$ 
\\ 
\midrule
T2V-DDPM \cite{T2v-ddpm} & 22.03 & \underline{0.759} & \underline{0.074}  & 63.40  & 14.72 & 0.645 & 0.249 &\underline{137.65} 
 & 11.35 & 0.373  & 0.421  &144.55 
 &  23.75 & \textbf{0.883} & \textbf{0.050} & \underline{23.88} & 14.97 & \underline{0.835} & 0.122 & 54.09 & 11.33 & 0.502  & 0.364  &138.90 
\\ 
CycleGAN \cite{CycleGAN} & 19.58 & 0.521 & 0.185  & 45.34  & \underline{17.29} & \underline{0.671} & \underline{0.223} &144.46 
 & 15.02  & \underline{0.519}  & 0.382  &122.77 
 &   18.58 & 0.452 & 0.272  & 87.13 & 19.03 & 0.740 & 0.204 & 106.65 
& 14.42 & 0.396  & 0.378  &113.75 
\\ 
UNIT \cite{UNIT} & 23.02 & 0.702 & 0.140  & 59.69    & 12.51 & 0.107 & 0.414 &176.83 
 & 13.96  & 0.421  & 0.396   &136.60 
 &   \underline{25.03} & 0.744 & 0.138  & 50.20 & 13.86 
& 0.194 & 0.405 & 175.24 
& 16.86  & 0.577  & 0.349  &117.75 
\\ 
MUNIT \cite{MUNIT} & \underline{23.34} & 0.694 & 0.141  & 48.37   & 14.74 & 0.402 & 0.318 &163.14 
 & 13.06 & 0.370  & 0.450  &144.65  
 &   23.56 & 0.664 & 0.173  & 63.66  & 19.25 
& 0.528 & \underline{0.104} & \underline{51.76} 
& 15.86  & 0.549  & 0.374 &126.22 
\\
DCLGAN \cite{DCLGAN} & 18.08 & 0.492 & 0.291  & 80.46   & 14.73 & 0.502 & 0.297 &176.15 
 & 13.62 & 0.364 & 0.431   &119.28 
 &   17.31 & 0.424 & 0.333  & 93.57  & 18.11 
& 0.493 & 0.282 & 196.44 
& 12.99  & 0.440  & 0.474  &191.5 
\\
I2V-GAN \cite{I2v-gan} & 15.23 & 0.403 & 0.378 & 134.97   & 14.31 & 0.518 & 0.321 &202.39  
 & 12.47 & 0.391  & 0.447  &179.09  
 &   15.98 & 0.443 & 0.398 & 147.49  & 19.33 
& 0.664 & 0.230 & 143.94 
& 11.31 & 0.382  & 0.445  &143.67 
\\
ROMA \cite{ROMA} & 19.10 & 0.531 & 0.173 & \underline{33.47}  & 12.88 & 0.128 & 0.629 &250.61 
 & \textbf{17.37}  & 0.464  & 0.293  &\underline{83.46}
 &   19.83 & 0.541 & 0.193 & 40.36  & \underline{20.20} & 0.613 & 0.174 & 83.17
& \underline{19.00}  & \underline{0.606}  & \underline{0.251}  &\underline{75.96} 
\\
CUT \cite{CUT} & 14.14 & 0.338 & 0.460 & 122.71   & 16.87 & 0.535 & 0.272 &142.45 
 & 15.21  & 0.401  & \underline{0.371}  &104.66 
 &   18.78 & 0.486 & 0.270 & 76.74  & 15.84 & 0.561 & 0.359 & 228.34 
& 9.33  & 0.231  & 0.631  &314.04 
\\
Ours& \textbf{23.67} & \textbf{0.835} & \textbf{0.054} & \textbf{22.80} & \textbf{19.89} & \textbf{0.774} & \textbf{0.185} &\textbf{94.11} 
 & \underline{17.03}  & \textbf{0.554} & \textbf{0.184}  &\textbf{66.19} 
 &   \textbf{27.96} & \underline{0.871} & \underline{0.054}  & \textbf{21.33}  & \textbf{24.03} & \textbf{0.883} & \textbf{0.096} & \textbf{46.18}  & \textbf{23.54}  & \textbf{0.765}  & \textbf{0.116}  &\textbf{57.26} \\
 \bottomrule
\end{tabular}
}
\end{table*}

\paragraph{(3) Constraint Loss Function for Statistical Regulation.}
To implement the conditional term $p(w \mid \tilde{\mathbf{x}}_0)$ in our SCI framework, we define two differentiable losses—\emph{channel constraint loss} $\mathcal{L}_{ccl}$ and \emph{statistical constraint loss} $\mathcal{L}_{scl}$—which correspond to $\mathcal{K}(\cdot)$ and $\mathcal{Q}(\cdot)$, respectively. These are combined into the overall constraint objective:
\begin{equation}
\mathcal{L}_{\mathrm{cons}} = \lambda_{ccl}\cdot\mathcal{L}_{ccl} \;+\; \lambda_{scl}\cdot\mathcal{L}_{scl}.
\end{equation}

\textbf{Channel Constraint Loss.} We penalize discrepancies in per-channel histograms between the noisy estimate $\tilde{\mathbf{x}}_0$ and the empirical prior histograms from the target modality:
\begin{equation}
\mathcal{L}_{ccl} = \sum_{\delta \in \{R,G,B\}} \sum_{i=1}^{B} \frac{\bigl(h_{i}^{\text{pred},\delta} - h_{i}^{\text{prior},\delta}\bigr)^{2}}{h_{i}^{\text{pred},\delta} + h_{i}^{\text{prior},\delta} + \epsilon},
\end{equation}
where $h_{i}^{\text{pred},\delta}$ and $h_{i}^{\text{prior},\delta}$ are the $i$-th histogram bin counts of the predicted and prior distributions for channel $\delta$, $B$ is the number of bins, and $\epsilon\!=\!10^{-6}$ avoids division by zero.

\textbf{Statistical Constraint Loss.} To align the generated image’s first- and second-order channel statistics with the target domain, we define:
\begin{equation}
\mathcal{L}_{scl} = \sum_{\delta \in \{R,G,B\}} \bigl|\mu_{\delta}^{\text{pred}} - \mu_{\delta}^{\text{prior}}\bigr|
\;+\;\bigl|\sigma_{\delta}^{\text{pred}} - \sigma_{\delta}^{\text{prior}}\bigr|,
\end{equation}
where $\mu_{\delta}^{\text{pred}}, \sigma_{\delta}^{\text{pred}}$ are the mean and standard deviation of channel $\delta$ in $\tilde{\mathbf{x}}_0$, and $\mu_{\delta}^{\text{prior}}, \sigma_{\delta}^{\text{prior}}$ are the corresponding empirical values from the training data.

By backpropagating the gradients of $\mathcal{L}_{\mathrm{cons}}$ into the reverse sampling step, we adjust the denoising mean via
\[
\mu \;\leftarrow\; \mu \;+\; \Sigma \cdot\nabla_{\tilde{\mathbf{x}}_0}\mathcal{L}_{\mathrm{cons}},
\]
thereby enforcing both histogram consistency and statistical alignment without modifying the original diffusion model parameters.

The overall inference process using the SCI is presented in Algorithm \ref{Algorithm2}.
\begin{algorithm}[t]
\renewcommand{\algorithmicrequire}{\textbf{Input:}}
\renewcommand{\algorithmicensure}{\textbf{Output:}}
	\caption{CM-Diff Inference (Visible-to-Infrared)}
    \label{Algorithm2}
    \begin{algorithmic}[1]
        \REQUIRE Visible image  $\mathbf{x}_{0}^{\textsc{vis}}$, infrared-to-visible direction label $\mathcal{C}^{\textsc{ir}}$.
        \ENSURE Translated IR modality image  $\mathbf{\hat{x}}_{0}^{\textsc{ir}}$
        \STATE Initialize a pre-trained edge detection network DexiNed\\
        \STATE $\mathbf{x}_{0}^{\textsc{vis}} \in \mathcal{X}^{\textsc{vis}}$ , $\mathbf{\hat{x}}_{T}^{\textsc{ir}} \sim \mathcal{N} \left( \mathbf{0}, \mathbf{I} \right)$
        \STATE $E^{\textsc{vis}} \gets \text{DexiNed}\left(\mathbf{x}_{0}^{\textsc{vis}}\right)$
        \STATE $J^{\textsc{ir}} \gets \tau_{\phi }(\text{$\mathcal{C}^{\textsc{ir}}$})$
          \FOR{$t=T,...,1$}
            \STATE  $\mu, \Sigma=\mu_{\phi }\left(\mathbf{\hat{x}}_{t}^{\textsc{ir}}\right), \Sigma_{\phi }\left(\mathbf{\hat{x}}_{t}^{\textsc{ir}}\right)$
            \STATE  $\tilde{\mathbf{x}}_{0}^{\textsc{ir}}=\frac{\mathbf{\hat{x}}_{t}^{\textsc{ir}}}{\sqrt{\bar{\alpha}_{t}}}-\frac{\sqrt{1-\bar{\alpha}_{t}} \cdot \epsilon_{\phi }\left(\mathbf{\hat{x}}_{t}^{\textsc{ir}},\mathbf{x}_{0}^{\textsc{vis}},E^{\textsc{vis}},J^{\textsc{ir}},t\right)}{\sqrt{\bar{\alpha}_{t}}}$  
            \STATE  $\mathcal{L}_{{cons}} = \lambda_{scl}\cdot \mathcal{L}_{scl}\left ( \tilde{\mathbf{x}}_{0}^{\textsc{ir}}, w \right ) + \lambda_{ccl} \cdot \mathcal{L}_{ccl}\left ( \tilde{\mathbf{x}}_{0}^{\textsc{ir}} \right ) $
            \STATE  Sample $\mathbf{\hat{x}}_{t-1}^{\textsc{ir}}$ by $\mathcal{N}\left ( \mu+ \nabla_{\tilde{\mathbf{x}}_{0}^{\textsc{ir}}}\mathcal{L}_{{cons}},\Sigma \right ) $
        \ENDFOR
        \STATE return $\mathbf{\hat{x}}_{0}^{\textsc{ir}}$
    \end{algorithmic}
\end{algorithm}
\begin{figure*}[h]
    \centering
    \includegraphics[scale=0.61, trim=0mm 185mm 0mm 165mm, clip, width=1\textwidth]{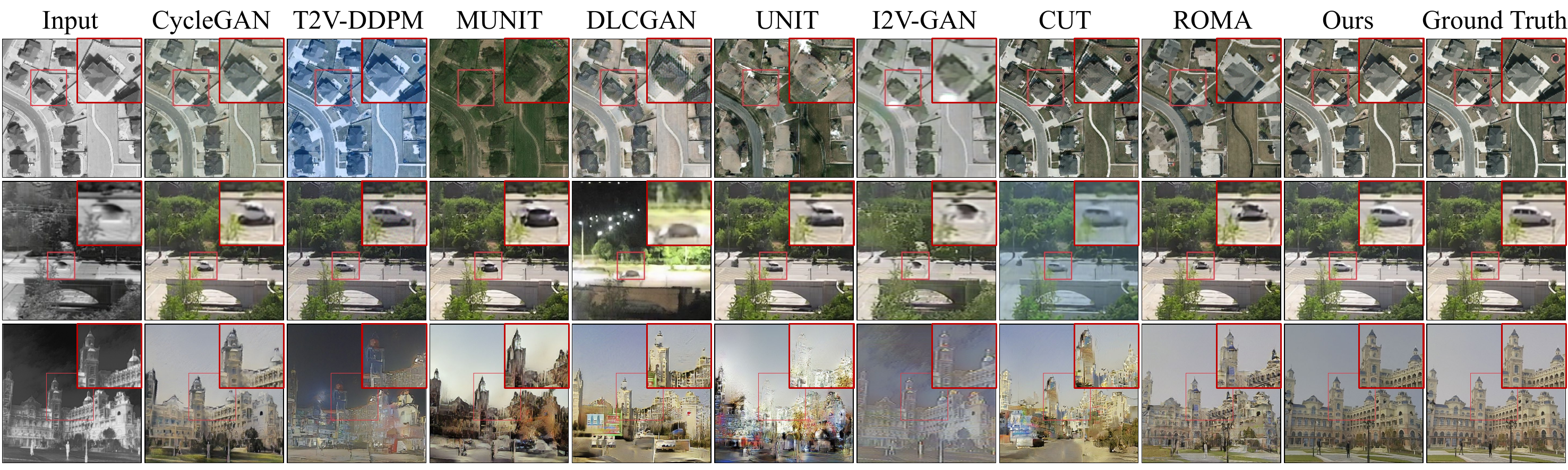}
    \caption{Qualitative results of CM-Diff for the infrared-to-visible image translation task. The first row corresponds to the VEDAI dataset, the second row to the AVIID dataset, and the third row to the $\text{M}^{3}\text{FD}$ dataset.}
    \label{figure4}
\end{figure*} 
\section{Experiments}\label{sec:ExAndRe} 
We conduct all our experiments on VEDAI \cite{VEDAI}, AVIID \cite{AVIID}, and  $\text{M}^{3}\text{FD}$ \cite{M3FDdataset} datasets to evaluate the performance of image translation methods. 
\begin{figure*}[h]
    \centering
    \includegraphics[scale=0.61, trim=0mm 185mm 0mm 165mm, clip, width=1\textwidth]{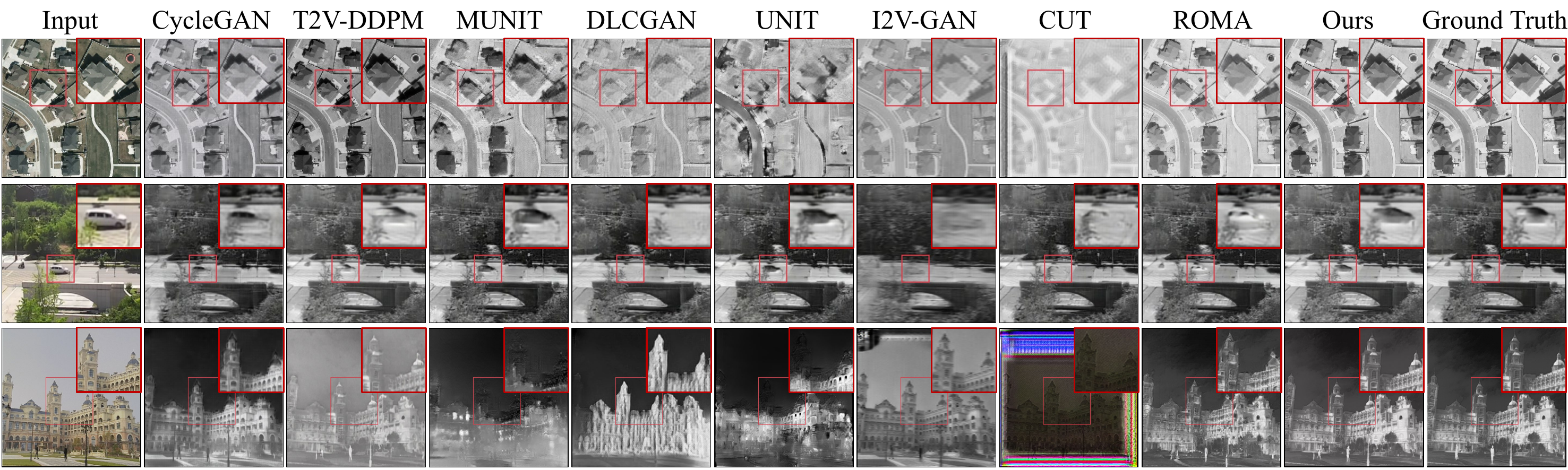}
    \caption{Qualitative results of CM-Diff for the visible-to-infrared image translation task. The first row corresponds to the VEDAI dataset, the second row to the AVIID dataset, and the third row to the $\text{M}^{3}\text{FD}$ dataset.}
    \label{figure5}
\end{figure*} 
\subsection{Comparison with State-of-the-Art Methods}
\paragraph{(1) Quantitative Results.} 
The quantitative comparison results are listed in Tab. \ref{table1}.
The evaluation metrics include FID \cite{FID}, SSIM \cite{SSIM}, LPIPS \cite{LPIPS}, and PSNR for both the infrared-to-visible image translation task and the visible-to-infrared image translation task.
Overall, our method demonstrates superior performance compared to unidirectional and bidirectional modality translation methods.

Compared to the DDPM-based method T2V-DDPM \cite{T2v-ddpm}, the proposed method achieves a 7.44\% improvement in PSNR, along with a 64.04\% reduction in FID for the infrared-to-visible image translation task on the AVIID  dataset. 
On the $\text{M}^{3}\text{FD}$  dataset, the enhancements are also pronounced, with a 53.04\% increase in PSNR and a 56.29\% reduction in LPIPS.
This improvement stems from the limitations of diffusion-based unidirectional methods, which focus solely on learning the data distribution in the target modality and establishing mappings within it, thereby missing deeper semantic connections. In contrast,  BDT can simultaneously learn the data distributions of both modalities. 
Additionally, the CFC is utilized to enhance the semantic guidance of the source modality image.

Compared to GAN-based unidirectional translation methods, our proposed method achieves improvements across the three datasets. 
On the AVIID  dataset, for the infrared-to-visible image translation task, FID decreases by 72.20\% and PSNR increases by 48.88\% compared to CUT \cite{CUT}, while FID decreases by 31.88\% and PSNR increases by 23.93\% compared to ROMA \cite{ROMA}. 
This improvement can be attributed to the proposed DDPM-based method, which transforms the data distribution into a series of noise distributions to avoid the problem of mode collapse. Additionally, it employs SCI during the sampling phase to incorporate observed statistical information, thereby improving the consistency between the generated results and the real data distribution.

Compared to GAN-based bidirectional translation methods, our proposed method also demonstrates performance gains. On the AVIID  dataset, for the infrared-to-visible image translation task, PSNR increases by 1.41\% compared to MUNIT \cite{MUNIT}.  
The observed improvement can be ascribed to the adoption of a BDT, which effectively encodes directional embeddings and edge information while leveraging multi-scale features extracted by the source modality encoder. This method mitigates modality distribution ambiguity arising from the structural and semantic similarities between infrared and visible images.

\paragraph{(2) Qualitative Results.} 
We compare the infrared-to-visible (Fig. \ref{figure4}) and visible-to-infrared (Fig. \ref{figure5}) translation results of state-of-the-art methods with our approach. In the first row, using the VEDAI  dataset, we observe that all methods transfer semantic information well, but our method outperforms DCLGAN \cite{DCLGAN} and UNIT \cite{UNIT} in terms of color and detail. This is due to the limitations of DCLGAN and UNIT in accurately defining object boundaries and utilizing spatial information.
In the second row, methods like CycleGAN \cite{CycleGAN}, MUNIT \cite{MUNIT}, and ROMA \cite{ROMA} produce blurred or distorted objects. While CUT \cite{CUT}, T2V-DDPM \cite{T2v-ddpm}, and our method generate clearer object structures, our method stands out by improving color fidelity through inference guidance with the SCI.
The third row shows challenging samples with complex backgrounds and pedestrians. Methods like T2V-DDPM and CUT tend to mistakenly render background buildings as sky. In contrast, our method accurately reconstructs architectural details and sky backgrounds, thanks to the CFC module that ensures precise feature generation and noise control.
\subsection{Ablation Study}
\paragraph{(1) Effect of CFC.} 
We conducted ablation experiments to demonstrate the effectiveness of the source modality features and edge feature map cues in preserving content consistency in translation results.
The quantitative results are presented in Tab. \ref{table4}.
It can be observed that the removal of the modality encoders and the edge detection part leads to a notable drop in the overall performance in both translation directions. 
We can see that external control using modality and edge features can effectively help the network distinguish boundaries of objects in the image. 
\begin{table}[th]
    \caption[]{Effect of CFC.} \label{table4}
    \centering
    \resizebox{0.48\textwidth}{!}{ 
        \begin{tabular}{c|cccc|cccc}
        \toprule
     \multirow{2}{*}{CFC}& \multicolumn{4}{c|}{infrared $\rightarrow$ visible} & \multicolumn{4}{c}{visible $\rightarrow$ infrared
}\\
         ~ & PSNR$\uparrow$ & SSIM$\uparrow$ & LPIPS$\downarrow$ & FID$\downarrow$  & PSNR$\uparrow$ & SSIM$\uparrow$ & LPIPS$\downarrow$ &FID$\downarrow$  
\\ 
        \midrule
         w/o  & 19.17 & 0.761 & 0.201  & 101.13   & 24.05 & 0.880 & 0.098 &47.84  \\ 
        w/ & \textbf{19.89} & \textbf{0.774} & \textbf{0.185} & \textbf{94.11}   & \textbf{24.06}  & \textbf{0.883} & \textbf{0.096} &\textbf{46.18}  
\\
    \bottomrule
    \end{tabular}
    }
\end{table}

\paragraph{(2) Effect of TDG.} 
In Tab. \ref{table7}, we investigate the impact of TDG on the quality of translation results for the two directions. 
When direction labels are passed through the label embedding layer and used to guide the network, the PSNR and SSIM metrics for the infrared-to-visible image translation task improve by 7.16\% and 6.91\%, respectively. 
Similarly, for the visible-to-infrared translation task, these metrics increase by 5.29\% and 14.52\%, respectively. 
These results demonstrate that TDG is indispensable during training, as it helps the model recognize noisy data and provide crucial guidance. 
Moreover, during inference, the TDG corresponds to assisting the model in executing the correct translation direction, further highlighting its importance.
\begin{table}[th]
    \caption[]{Effect of TDG.} \label{table7}
    \centering
    \resizebox{0.48\textwidth}{!}{ 
        \begin{tabular}{c|cccc|cccc}
        \toprule
     \multirow{2}{*}{TDG}& \multicolumn{4}{c|}{infrared $\rightarrow$ visible} & \multicolumn{4}{c}{visible $\rightarrow$ infrared
}\\
         ~ & PSNR$\uparrow$ & SSIM$\uparrow$ & LPIPS$\downarrow$ & FID$\downarrow$  & PSNR$\uparrow$ & SSIM$\uparrow$ & LPIPS$\downarrow$ &FID$\downarrow$  
\\ 
        \midrule
         w/o  & 18.56& 0.724& 0.253& 142.79 & 22.85& 0.771& 0.181&163.87 \\ 
        w/ & \textbf{19.89} & \textbf{0.774} & \textbf{0.185} & \textbf{94.11}   & \textbf{24.06}  & \textbf{0.883} & \textbf{0.096} &\textbf{46.18}  
\\
    \bottomrule
    \end{tabular}
    
    }
\end{table}

\paragraph{(3) Effect of SCI.} 
As listed in Tab. \ref{table5}, in the infrared-to-visible image translation task, the $\mathcal{L}_{scl}$ plays a significant role in enhancing modality translation performance, resulting in an improvement of 5.85\% in PSNR and 1.57\% in SSIM compared to the case without SCI.  
In the visible-to-infrared image translation task, it results in an improvement of 52.70\% in PSNR, 5.87\% in SSIM, 26.15\% in LPIPS, and 19.97\% in FID.  
Moreover, in the visible-to-infrared image translation task, incorporating the $\mathcal{L}_{ccl}$ further boosts the model's performance, leading to an improvement of 0.16\% in PSNR and 0.21\% in FID.
\begin{table}[th]
    \caption[]{Effect of SCI. } \label{table5}
    \centering
    \resizebox{0.48\textwidth}{!}{ 
        \begin{tabular}{cc|cccc|cccc}
        \toprule
     \multirow{2}{*}{$\mathcal{L}_{{ccl}}$} &  \multirow{2}{*}{$\mathcal{L}_{{scl}}$} & \multicolumn{4}{c|}{infrared $\rightarrow$ visible} & \multicolumn{4}{c}{visible $\rightarrow$ infrared
}\\
         ~ & ~ & PSNR$\uparrow$ & SSIM$\uparrow$ & LPIPS$\downarrow$ & FID$\downarrow$  & PSNR$\uparrow$ & SSIM$\uparrow$ & LPIPS$\downarrow$ &FID$\downarrow$  
\\ 
        \midrule
         ×  & × & 18.78 & 0.762 & 0.205 & 105.02   & 15.73 & 0.834 & 0.130 &57.83   \\ 
        ×  & \checkmark & 19.88 & 0.774 & \textbf{0.185}  & \textbf{93.65}    & 24.02 & \textbf{0.883} & \textbf{0.096} &46.28    
\\ 
        \checkmark  & × & 19.54 & \textbf{0.775} & 0.201 & 97.39   & 15.64 & 0.840 & 0.129 &56.88   
\\ 
        \checkmark & \checkmark & \textbf{19.89} & 0.774 & \textbf{0.185} & 94.11   & \textbf{24.06} & \textbf{0.883} & \textbf{0.096} &\textbf{46.18}  
\\
        \bottomrule
    \end{tabular}
    }
\end{table}

\section{Conclusion}\label{sec:Conclusion} 
This paper introduces CM-Diff, a unified framework that utilizes DDPM for bidirectional translation between infrared and visible images.
Our approach leverages the diffusion model and incorporates a Bidirectional Diffusion Training (BDT) to distinguish and learn modality-specific data distributions while establishing robust bidirectional mappings.
We have also proposed a Statistical Constraint Inference (SCI) for maintaining consistency between the statistical distributions of the translated images and training images.
Extensive experiments have shown the effectiveness of our method.

\newpage
\twocolumn[\section*{CM-Diff: A Single Generative Network for Bidirectional Cross-Modality Translation Diffusion Model Between Infrared and Visible Images (Supplementary Material)}]  
\thispagestyle{empty}
\appendix
\begin{figure*}[ht]
    \centering
    \includegraphics[scale=0.60, trim=0mm 0mm 0mm 0mm, clip, width=\textwidth]{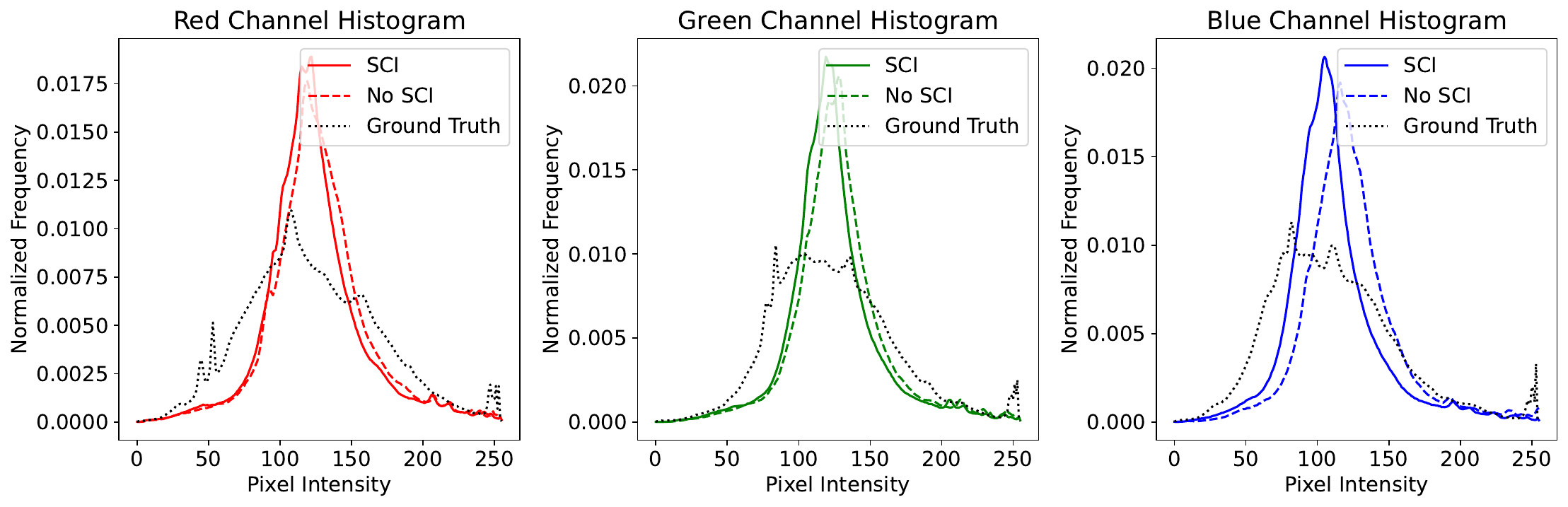}
    \caption{Comparison of histograms for each channel of visible images. The distribution of the images obtained using SCI, without SCI, and the ground truth visible images are presented. }
    \label{figure9}
\end{figure*} 
\begin{figure}[h]
    \centering
    \includegraphics[scale=0.60, trim=0mm 0mm 0mm 0mm, clip, width=0.485\textwidth]{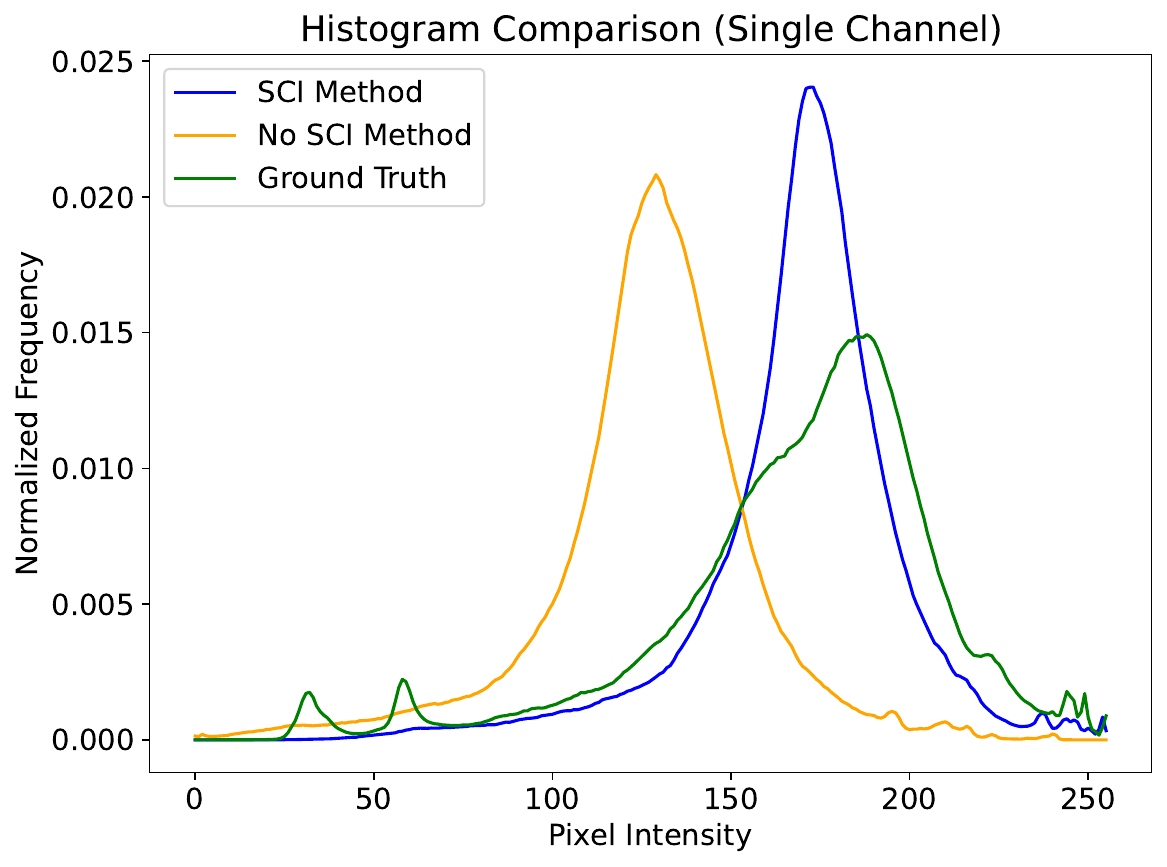}
    \caption{Comparison of histograms for infrared images. The distribution of the images obtained using SCI, without SCI, and the ground truth visible images are presented. }
    \label{figure10}
\end{figure} 
\textbf{Section A} evaluates the effectiveness of the proposed Statistical Constraint Inference (SCI) strategy through pixel intensity frequency histogram analysis. The results demonstrate that SCI enhances infrared-to-visible translation by effectively suppressing unnatural color artifacts and aligning the translated image distributions with those of the ground truth. Details of the CM-Diff architecture are also provided, including the Translation Direction Guidance (TDG), Cross-Modality Feature Control (CFC), and SCI modules.

\textbf{Section B} presents a comprehensive performance evaluation of cross-modality translation methods on object detection tasks using CFINet on the VEDAI dataset. Experimental results indicate that CM-Diff consistently outperforms state-of-the-art approaches such as ROMA and CUT in both infrared-to-visible and visible-to-infrared directions, achieving superior mAP scores. Moreover, translation experiments conducted on the Rain100H dataset further validate the model’s generalization capability beyond the IR-VIS domain.

\textbf{Section C} presents ablation and comparison studies that validate the effectiveness of key components in CM-Diff. Varying the constraint scale factors $\lambda_{\text{ccl}}$ and $\lambda_{\text{scl}}$ shows that moderate values (e.g., 20) offer the best balance between fidelity and diversity. Among the tested histogram distance metrics, $\chi^2$ achieves the lowest FID and best perceptual quality. For structural guidance, DexiNed outperforms traditional edge detectors by better preserving semantic contours. Finally, comparisons with general I2I methods (UGATIT, UNSB, DDBM) highlight CM-Diff’s superior performance in cross-modality alignment, confirming the importance of domain-specific design.

\textbf{Section D} provides qualitative results on the M$^3$FD, AVIID, and VEDAI datasets, illustrating the robustness of CM-Diff across a variety of challenging scenes. Analysis of training losses reveals that IR-to-VIS translation presents relatively lower optimization difficulty. Additional implementation details are also included, such as the use of a U-Net backbone with multi-scale attention, DexiNed-based edge fusion, and AdamW optimization. Evaluation metrics including FID, SSIM, LPIPS, and PSNR are discussed, along with detailed descriptions of the datasets used in the experiments.

\section{A. Method and Module Analysis}
\subsection{Overall Framework}
\begin{figure*}[ht]
    \centering
    \includegraphics[scale=0.60, trim=7mm 37mm 7mm 160mm, clip, width=\textwidth]{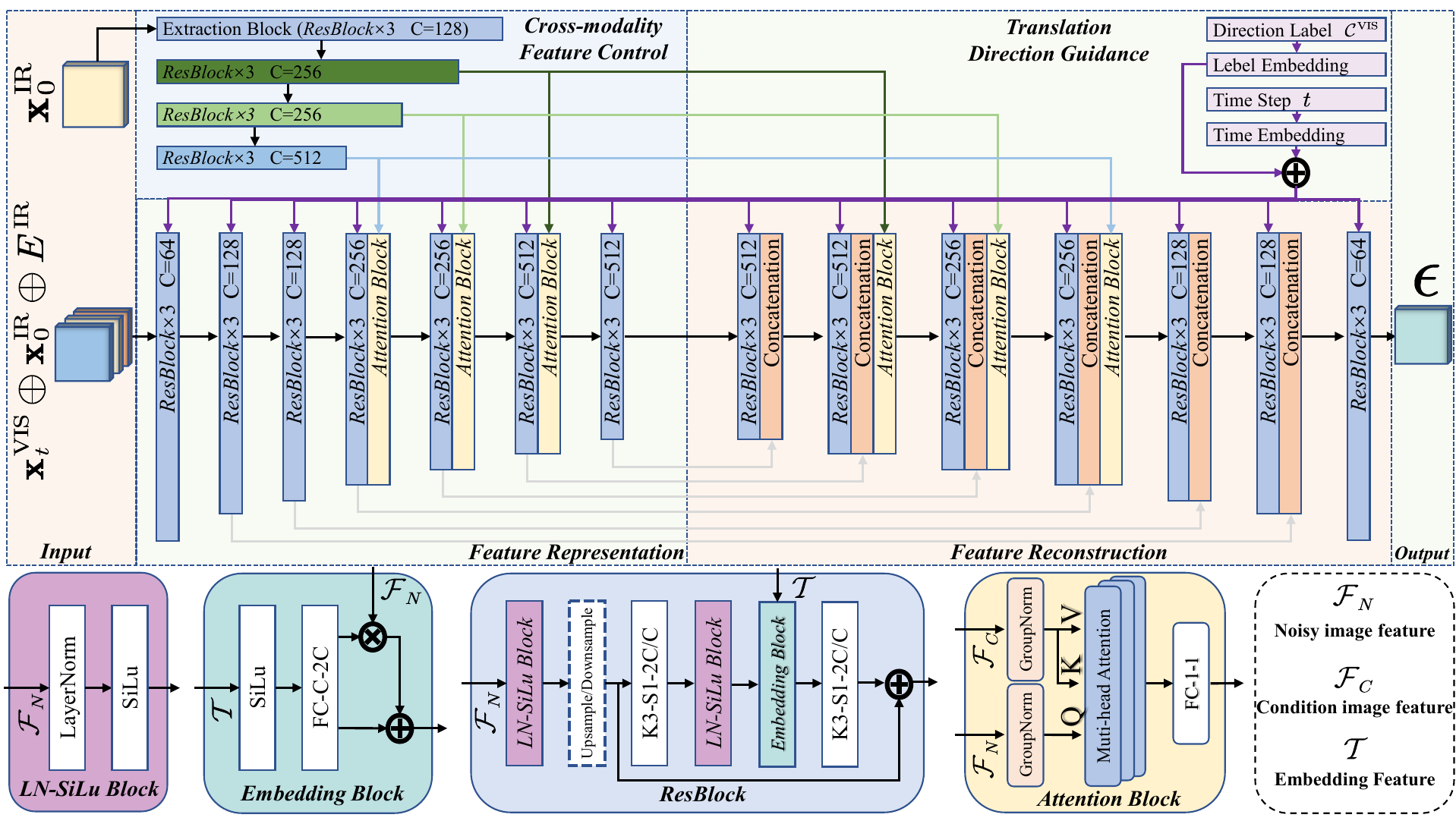}
    \caption{ {CM-Diff Architecture:} Comprising Cross-Modality Feature Control (CFC), Translation Direction Guidance (TDG), and the Feature Representation and Reconstruction Network. CM-Diff processing a noisy input image by integrating edge feature maps and conditional modality images. CFC aligning noise translation with the reverse direction of semantic content in the conditional modality. TDG guiding the predicted noise toward the expected modality's reverse direction. Finally, the network predicting the output noise.
}
    \label{figure:UNet}
\end{figure*}

Figure~\ref{figure:UNet} depicts the complete CM-Diff architecture, which comprises three core components:

\noindent\textbf{(1) Translation Direction Guidance (TDG)}: steers the diffusion noise prediction toward the reverse gradient of the target modality;

\noindent\textbf{(2) Cross-Modality Feature Control (CFC)}: injects semantic information across modalities at multiple feature scales to ensure content consistency;

\noindent\textbf{(3) Statistical Constraint Inference (SCI)}: applies histogram-based statistical constraints ($\mathcal{L}_{\mathrm{ccl}}$, $\mathcal{L}_{\mathrm{scl}}$) during sampling to reduce distribution shift.
\subsection{SCI Module Deep Dive and Analysis}
The SCI design draws inspiration from guided diffusion and distribution alignment techniques. By adjusting the mean and covariance of the target modality in the noise steps, it enforces closer adherence to the true data distribution. Histogram comparisons in Figures~\ref{figure9} and~\ref{figure10} confirm SCI’s efficacy for bidirectional infrared–visible translation.

As illustrated in Fig.~\ref{figure9}, the proposed Statistical Constraint Inference (SCI) strategy introduces significant improvements in the infrared-to-visible image translation task. Specifically, when employing SCI, the frequency distributions of pixel intensities for each RGB channel shift notably towards lower intensity values. This shift is particularly evident in the green and blue channels, where high-frequency components associated with unnatural intensity peaks are substantially suppressed. As a result, the occurrence of color artifacts—such as oversaturation or unnatural hues—is greatly reduced in the translated visible images.

The effectiveness of SCI is further validated in the reverse translation task, i.e., visible-to-infrared conversion. As shown in Fig.~\ref{figure10}, the translated infrared images exhibit histogram distributions that are more consistent with those of the ground truth infrared images when SCI is applied. This consistency reflects a more faithful preservation of modality-specific statistical properties during translation.

Overall, these observations highlight the advantage of integrating SCI into our framework. By enforcing statistical alignment between the translated output and the target domain, SCI not only improves the perceptual realism of the results but also mitigates the color distribution anomalies typically introduced by Bidirectional Diffusion Training (BDT). Consequently, the proposed \textbf{CM-Diff} model, when guided by SCI, demonstrates superior capability in performing high-fidelity, artifact-free cross-modality translation within a unified generative framework.

\section{B. Performance Evaluation and Expandability}
\subsection{Generated Data Efficacy in Object Detection}
\begin{table}
\caption{Quantitative evaluation of visible-to-infrared image translation performance of various translation methods based on CFINet.}
    \label{tab:2}
    \centering
    \resizebox{0.48\textwidth}{!}{
    \begin{tabular}{l|ccccc}
    \toprule
         Method&  P&  R&  $mAP_{50}$ &$mAP_{75}$& $mAP$\\
         \midrule
         ROMA \cite{ROMA}&  0.133
&  0.027&  0.036
 &0.022& 0.012
\\
         
CUT \cite{CUT}&  0.418&  0.157
&  0.187
 &0.159& 0.082
\\
         I2V-GAN \cite{I2v-gan} &  0.379&  0.113&  0.179 &0.144& 0.077
\\
 Ground Truth& 
0.471& 0.368& 0.449 &0.306&0.168
\\
         
Ours&  
0.313&  0.198&  0.281 &0.249& 0.138
\\
\bottomrule
    \end{tabular}
    }
\end{table}

\begin{table}
\caption{Quantitative evaluation of infrared-to-visible image translation performance of various translation methods based on CFINet.}
    \label{tab:3}
    \centering
    \resizebox{0.48\textwidth}{!}{
    \begin{tabular}{l|ccccc}
    \toprule
         Method&  P&  R&  $mAP_{50}$ &$mAP_{75}$& $mAP$\\
         \midrule
         ROMA \cite{ROMA}
&  0.240
&  0.106
&  0.175
&0.136& 0.072
\\
         
CUT \cite{CUT}&  0.185
&  0.025
&  0.068
&0.059& 0.032
\\
         I2V-GAN \cite{I2v-gan}&  0.168&  0.079&  0.129&0.090& 0.048
\\
 Ground Truth& 
0.237& 0.179& 0.386&0.217&0.121\\
         
Ours&  

0.241&  0.165&  0.265&0.206& 0.105
\\
\bottomrule
    \end{tabular}
    }
\end{table}
We employ the recent small object detection method, CFINet \cite{CFINet}, and train it separately on infrared and visible data using the VEDAI dataset. Various translation methods are utilized to perform infrared-to-visible and visible-to-infrared modality translation. The translation results are then evaluated using CFINet
. For comparison, we selected several high-performing modality translation methods, including ROMA \cite{ROMA}, CUT \cite{CUT}, and I2V-GAN \cite{I2v-gan}. The quantitative results for the infrared and visible modalities are shown in Tables \ref{tab:2} and \ref{tab:3}, respectively.

\textit{In the comparison of infrared modality translation results}, significant differences can be observed in terms of precision (P), recall (R), and mean average precision (mAP). As shown in the table, the Ground Truth achieves the best performance, with the highest precision (0.471) and recall (0.368). It also attains the highest mAP values, including mAP$_{50}$ = 0.449, mAP$_{75}$ = 0.306, and an overall mAP of 0.168.  
Among the modality translation methods, our approach demonstrates the best performance, achieving mAP$_{50}$, mAP$_{75}$, and overall mAP values of 0.281, 0.249, and 0.138, respectively. These results are significantly superior to those of ROMA, CUT, and I2V-GAN. Additionally, our method exhibits a recall of 0.198, which is substantially higher than that of other approaches, indicating its effectiveness in the object detection task.  
Regarding the other modality translation methods, CUT and I2V-GAN perform similarly in terms of precision and mAP, though CUT consistently achieves slightly better results. In contrast, ROMA exhibits the weakest performance across all metrics, suggesting that its translated infrared images contribute minimally to the detection task. In summary, our method generates higher-quality infrared images and substantially improves the performance of object detection tasks.  

\textit{In the comparison of visible modality translation results}, significant differences can be observed in terms of precision (P), recall (R), and mean average precision (mAP). As shown in the table, the Ground Truth achieves the best performance, with relatively high precision (0.237) and recall (0.179). It also attains the highest mAP values, including mAP$_{50}$ = 0.386, mAP$_{75}$ = 0.217, and an overall mAP of 0.121.  
Among the modality translation methods, our approach demonstrates superior performance compared to others. Specifically, it achieves mAP$_{50}$, mAP$_{75}$, and overall mAP values of 0.265, 0.206, and 0.105, respectively, significantly outperforming ROMA, CUT, and I2V-GAN. Additionally, our method exhibits a recall of 0.165, second only to the Ground Truth, indicating its effectiveness in the object detection task under the visible modality.  
Regarding other translation methods, ROMA performs relatively well in terms of precision and mAP but has a lower recall (0.106) compared to our model. CUT and I2V-GAN demonstrate weaker overall performance, with CUT exhibiting the lowest scores across all metrics, suggesting that its translated visible images are of lower quality and less beneficial for object detection tasks. In summary, our method generates higher-quality visible images and achieves superior performance in object detection tasks.  

Finally, we observe that the quality of our infrared translation results is superior to that of visible translation results, both in terms of translation quality and object detection task performance. This observation is also supported by the training loss depicted in Fig. \ref{figure6}, where the loss for visible-to-infrared translation is consistently lower than that for infrared-to-visible translation. This phenomenon can be attributed to the inherent characteristics of infrared and visible images. Specifically, visible images consist of three channels representing the intensity values of red, green, and blue, whereas infrared images contain only a single channel reflecting infrared radiation intensity.

\subsection{Applicability to Weather Translation task}
As shown in Tab. \ref{Weather Translation}. 
We directly applied our approach to a weather translation task without any architectural modification and achieved competitive performance on the Rain100H dataset \cite{yang2017deep}, which consists of 100 pairs of synthetic rainy and clean images commonly used for image deraining. This result demonstrates the generalization capability of our method beyond the IR–VIS domain.

\begin{table}
\caption{Comparison of different methods on the Rain100H weather translation task.}
    \label{Weather Translation}
    \centering
    \resizebox{0.48\textwidth}{!}{
    \begin{tabular}{l|cc|cc}
    \toprule
         \multirow{2}{*}{Method} & \multicolumn{2}{c|}{Rain$\rightarrow$ Sunny}   & \multicolumn{2}{c}{Sunny$\rightarrow$ Rain} \\
         ~ &  PSNR$\uparrow$ &  SSIM$\uparrow$ &  PSNR$\uparrow$ &SSIM$\uparrow$ \\
         \midrule
CM-Diff (ours)&  19.72&  0.771&  19.81&0.803\\
         SEMI \cite{Wei_2019_CVPR}&  16.56&  0.486&  /&/\\
 DA-CLIP\cite{luo2023controlling}& 
33.91& 0.926& /&/\\
\bottomrule
    \end{tabular}
    }
\end{table}

\section{C. Ablation and Comparison Studies}
\subsection{Analysis of the Constraint Scale Factor}
To evaluate the effectiveness of constraint-based guidance in our generative framework, we conducted a series of ablation experiments focusing on the constraint scale factors $\lambda_{\text{ccl}}$ and $\lambda_{\text{scl}}$. As presented in Table~\ref{table:1}, we systematically varied these parameters and assessed the corresponding impact on cross-modality translation performance using standard evaluation metrics including PSNR, SSIM, LPIPS, and FID.

The results demonstrate a clear performance trend: as the constraint scale factors increased, the network exhibited consistent improvements across all metrics. This indicates that stronger enforcement of content and structure consistency enhances the model's capacity to preserve semantic information and spatial alignment during translation. Specifically, setting $\lambda_{\text{ccl}} = \lambda_{\text{scl}} = 20$ yielded the best overall performance, achieving the highest PSNR score of 19.89 and LPIPS score of 0.185 in the infrared-to-visible translation task, as well as the highest SSIM score of 0.883 in the visible-to-infrared task.

However, further increasing the constraint values beyond this optimal point (e.g., $\lambda=40$ and $\lambda=60$) resulted in a slight degradation of performance. For instance, when increasing the scale from 40 to 60, the visible-to-infrared translation task experienced a marginal drop of 0.12\% in PSNR and 0.11\% in LPIPS. This decline is attributed to the over-constraint of the mean estimation in the diffusion sampling process, which likely suppresses the model’s flexibility in modeling local uncertainty and fine-grained variations. These findings suggest that while constraint-based guidance is beneficial, excessive enforcement may hinder the generative diversity required for high-fidelity modality translation.
\begin{table}[h]
    \caption[]{Effect of the Constraint Scale Factors. The best performance for each metric is highlighted in bold.} \label{table:1}
    \centering
     \resizebox{0.48\textwidth}{!}{
        \begin{tabular}{c|cccc|cccc}
        \toprule
     \multirow{2}{*}{$\lambda_{{ccl}}, \lambda_{{scl}}$}& \multicolumn{4}{c|}{infrared $\rightarrow$ visible} & \multicolumn{4}{c}{visible $\rightarrow$ infrared}\\
         ~ & PSNR$\uparrow$ & SSIM$\uparrow$ & LPIPS$\downarrow$ & FID$\downarrow$  & PSNR$\uparrow$ & SSIM$\uparrow$ & LPIPS$\downarrow$ & FID$\downarrow$  
\\
         \midrule
        0& 18.78& 0.762& 0.205&105.02 & 15.73& 0.834& 0.130& 57.83 
\\ 
        10 & 19.88& \textbf{0.775}& 0.186& \textbf{93.66} & 22.78& 0.879& 0.099& 47.18 
\\ 
        20  & \textbf{19.89}& 0.774& \textbf{0.185}& 94.11 & 24.06& \textbf{0.883}& 0.096& 46.18 
\\ 
        40  & 19.85& 0.774& \textbf{0.185}& 93.74 & \textbf{24.13}& \textbf{0.883}& \textbf{0.095}& 46.01 
\\ 
        60  & 19.84& 0.774& \textbf{0.185}& 93.68 & 24.10& \textbf{0.883}& 0.096& \textbf{45.98} \\
        \bottomrule
    \end{tabular}
    }
\end{table}
\subsection{Histogram Distance Selection}
To enforce statistical alignment between generated and target domains, we evaluate three histogram distance measures—Euclidean, Bhattacharyya, and $\chi^2$—within the Statistical Constraint Inference framework (Table~\ref{Histogram Distance}). All metrics achieve comparable PSNR and LPIPS for infrared→visible translation and similar SSIM for visible→infrared translation, indicating their baseline effectiveness in distribution matching. Notably:

\begin{itemize}
  \item \textbf{Euclidean distance} attains the highest PSNR (24.08 dB) and lowest LPIPS (0.095) for IR→VIS, but exhibits a higher FID (46.75), suggesting marginally inferior deep‐feature congruence.
  \item \textbf{Bhattacharyya distance} yields PSNR and LPIPS on par with Euclidean for IR→VIS and slightly improves FID to 46.71, yet offers no significant benefit for VIS→IR.
  \item \textbf{$\chi^2$ distance (proposed)} incurs only a minor PSNR (–0.05 dB) and LPIPS (+0.001) penalty relative to Euclidean in IR→VIS, while achieving the lowest FID scores in both directions (46.18 for IR→VIS and 94.11 for VIS→IR). This reflects its enhanced sensitivity to sparse but extreme discrepancies in pixel‐intensity histograms, yielding crisper and more artifact‐free translations.
\end{itemize}

Consequently, the $\chi^2$ metric provides the optimal compromise between pixel‐level fidelity and perceptual realism and is therefore adopted as the default histogram distance.  

\begin{table}
\caption{Performance comparison of different histogram distance metrics in SCI for bidirectional IR-VIS translation.}
    \label{Histogram Distance}
    \centering
    \resizebox{0.48\textwidth}{!}{
    \begin{tabular}{l|cccc|cccc}
    \toprule
         \multirow{2}{*}{Histogram Distance
}& \multicolumn{4}{c|}{infrared $\rightarrow$ visible} &   \multicolumn{4}{c}{visible $\rightarrow$ infrared} \\
         ~ &  PSNR$\uparrow$ &  SSIM$\uparrow$  & LPIPS$\downarrow$ &FID$\downarrow$ &  PSNR$\uparrow$ &SSIM$\uparrow$  & LPIPS$\downarrow$ &FID$\downarrow$ \\
         \midrule
Euclidean&  \textbf{24.08}&  0.882
& \textbf{0.095}&46.75
&  19.88
&\textbf{0.774}& \textbf{0.185}&94.43
\\
         Bhattacharyya
&  24.08
&  0.882
& 0.096
&46.71
&  19.88
&\textbf{0.774}& \textbf{0.185}&94.45
\\
 Chi2 (ours)
& 
24.03
& \textbf{0.883}& 0.096
&\textbf{46.18}& \textbf{19.89}&\textbf{0.774}& \textbf{0.185}&\textbf{94.11}\\
\bottomrule
    \end{tabular}
    }
\end{table}

\subsection{Edge Detector Selection and Justification}
Effective cross‐modality feature integration hinges on robust structural priors. Conventional detectors such as Sobel and Canny employ fixed convolutional kernels to infer gradients, but their performance deteriorates in low‐contrast or noisy infrared imagery. To overcome these limitations, we integrate DexiNed, a deep learning–based edge extractor optimized for semantic contour delineation and fine‐grained detail preservation across heterogeneous visual conditions.

As shown in Table~\ref{Edge Detector}, DexiNed consistently outperforms both Sobel and Canny for bidirectional translation tasks. In the infrared→visible direction, DexiNed attains a superior PSNR of 24.03 dB and SSIM of 0.883, alongside the lowest LPIPS (0.096) and FID (46.18). Similarly, for visible→infrared translation, DexiNed achieves the highest PSNR (19.89 dB) and SSIM (0.774) while delivering competitive LPIPS and FID metrics. These quantitative improvements corroborate that DexiNed’s learned edge representations furnish more reliable and semantically meaningful guidance, thereby enhancing both pixel‐level accuracy and perceptual fidelity without incurring substantial computational overhead.

\begin{table}
\caption{Comparison of edge detectors for bidirectional IR-VIS translation.}
    \label{Edge Detector}
    \centering
    \resizebox{0.48\textwidth}{!}{
    \begin{tabular}{l|cccc|cccc}
    \toprule
         \multirow{2}{*}{Edge Detector
}& \multicolumn{4}{c|}{infrared $\rightarrow$ visible} &   \multicolumn{4}{c}{visible $\rightarrow$ infrared} \\
         ~ &  PSNR$\uparrow$ &  SSIM$\uparrow$  & LPIPS$\downarrow$ &FID$\downarrow$ &  PSNR$\uparrow$ &SSIM$\uparrow$  & LPIPS$\downarrow$ &FID$\downarrow$ \\
         \midrule
Sobel
&  23.64
&  0.774
& 0.109
&54.88
&  16.90
&0.563
& \textbf{0.180}&\textbf{63.14}\\
         Canny
&  21.70
&  0.729
& 0.142
&64.11
&  16.51
&0.537
& 0.199
&67.77
\\
 DexiNed (ours)& 
\textbf{24.03}& \textbf{0.883}& \textbf{0.096}&\textbf{46.18}& \textbf{19.89}&\textbf{0.774}& 0.185
&94.11
\\
\bottomrule
    \end{tabular}
    }
\end{table}
\begin{figure*}[ht]
    \centering
    \includegraphics[scale=0.60, trim=7mm 100mm 10mm 120mm, clip, width=\textwidth]{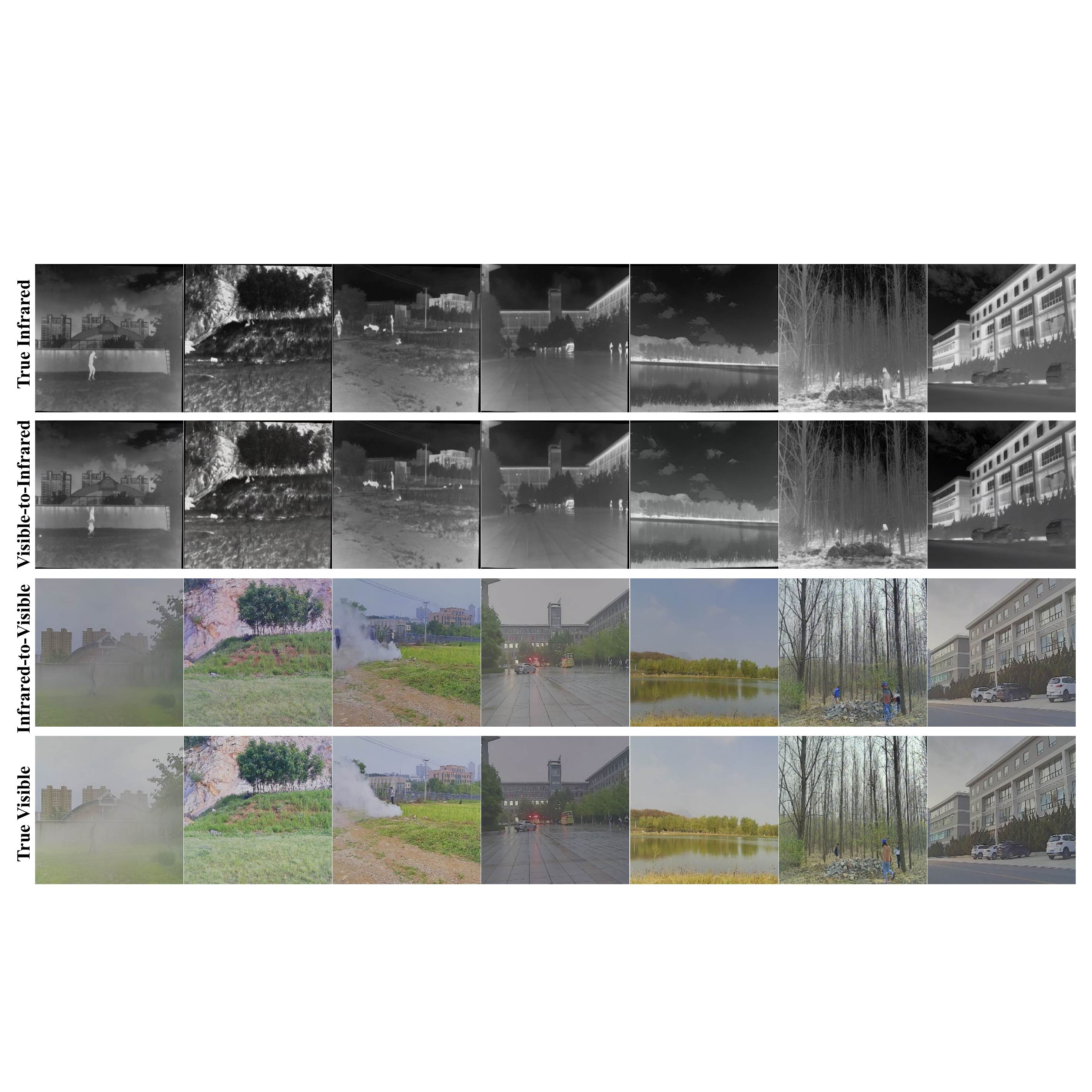}
    \caption{Modality translation results on the $\text{M}^{3}\text{FD}$ dataset. The first row represents real infrared images, the second and third rows show generated infrared and visible images, respectively, and the fourth row represents real visible images. }
    \label{figure:2}
\end{figure*} 
\subsection{Positioning Against General Image-to-Image Translation Methods}
General-purpose I2I models, optimized for RGB–to–RGB mappings, are ill-equipped to reconcile the substantial spectral and statistical disparities inherent in IR–VIS translation. As reported in Table~\ref{Performances}, UGATIT \cite{kim2019u} and UNSB \cite{kim2023unpaired} achieve only PSNR = 17.86 dB / 11.99 dB and SSIM = 0.602 / 0.415 for IR→VIS, with LPIPS = 0.358 / 0.433 and FID = 149.46 / 128.56, respectively—results that reflect both poor structural fidelity and perceptual realism. Even DDBM \cite{zhou2023denoising}, despite delivering a high PSNR of 26.26 dB, suffers from an excessively large FID of 227.34, indicating severe deep‐feature mismatches. By contrast, CM-Diff attains a well‐balanced performance (PSNR = 23.54 dB, SSIM = 0.765, LPIPS = 0.116, FID = 57.26 for IR→VIS), and maintains competitive metrics for VIS→IR (PSNR = 17.03 dB, SSIM = 0.554, LPIPS = 0.184, FID = 66.19). These quantitative contrasts confirm that our modality-specific bidirectional diffusion framework and statistical constraint inference are critical to bridging the IR–VIS domain gap. Consequently, we exclude general I2I baselines from the main text to preserve focus on specialized IR–VIS comparisons, relegating their evaluation to this supplementary section.  
\begin{table}
\caption{Performances of Different Methods on VEDAI.}
    \label{Performances}
    \centering
    \resizebox{0.48\textwidth}{!}{
    \begin{tabular}{l|cccc|cccc}
    \toprule
         \multirow{2}{*}{Histogram Distance}& \multicolumn{4}{c|}{infrared $\rightarrow$ visible} &   \multicolumn{4}{c}{visible $\rightarrow$ infrared} \\
         ~ &  PSNR$\uparrow$ &  SSIM$\uparrow$  & LPIPS$\downarrow$ &FID$\downarrow$ &  PSNR$\uparrow$ &SSIM$\uparrow$  & LPIPS$\downarrow$ &FID$\downarrow$ \\
         \midrule
UGATIT (ICLR’20)
&  17.86
&  0.602
& 0.358
&149.46
&  16.18
&0.461
& 0.379
&133.98
\\
         UNSB (ICLR’24)
&  11.99
&  0.415
& 0.433
&128.56
&  14.71
&0.408
& 0.475
&148.98
\\
 DDBM (ICLR’24)
& 
\textbf{26.26}& 0.748
& 0.405
&227.34
& \textbf{22.26}&\textbf{0.567}& 0.526
&241.35
\\
 CM-Diff (ours)
& 23.54
& \textbf{0.765}& \textbf{0.116}& \textbf{57.26}& 17.03
& 0.554
& \textbf{0.184}&\textbf{66.19}\\
\bottomrule
    \end{tabular}
    }
\end{table}
\section{D. Qualitative Results and Experimental Setup}
\subsection{Additional Translation Examples}
\begin{figure*}[ht]
    \centering
    \includegraphics[scale=0.60, trim=7mm 100mm 10mm 120mm, clip, width=\textwidth]{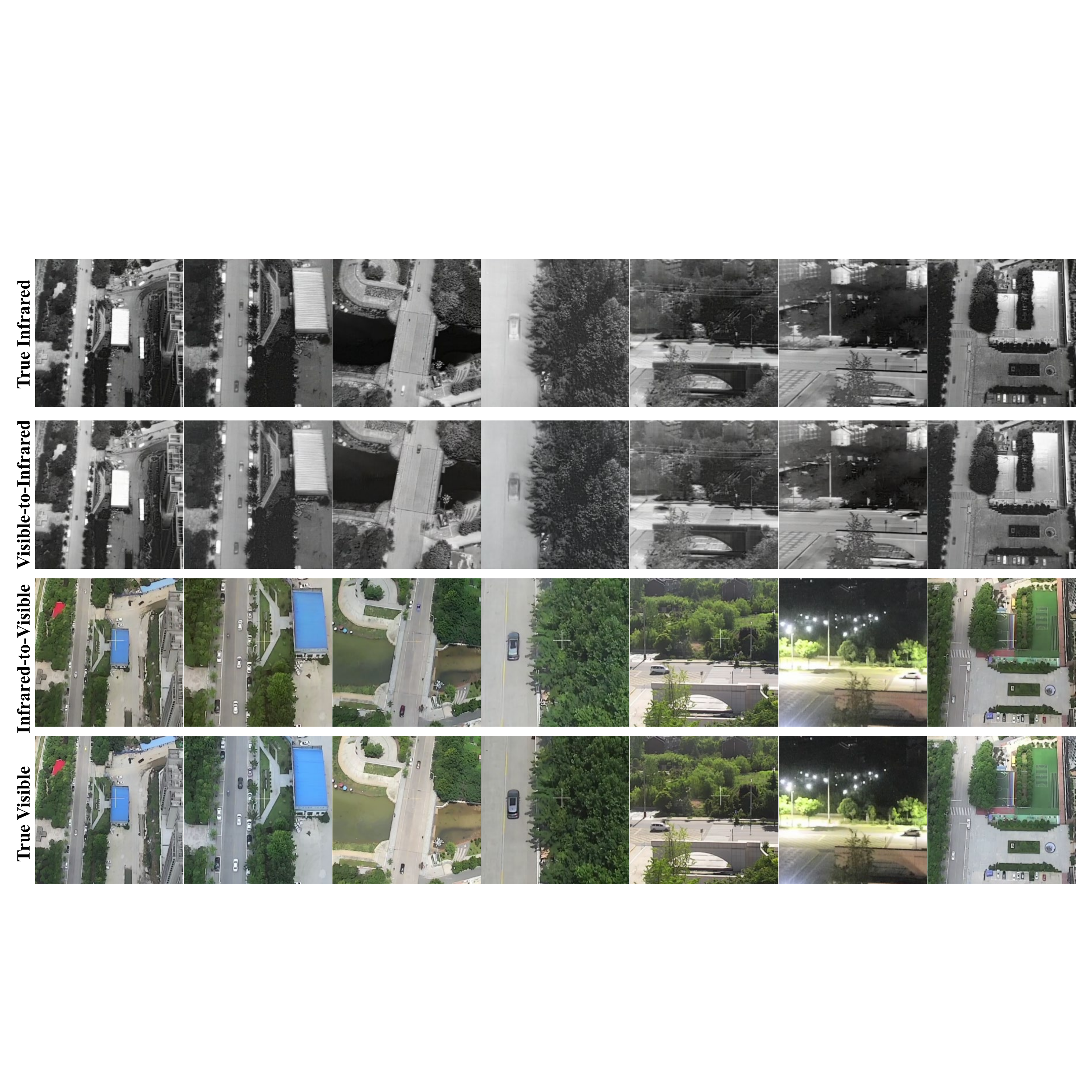}
    \caption{Modality translation results on the AVIID dataset. The first row represents real infrared images, the second and third rows show generated infrared and visible images, respectively, and the fourth row represents real visible images.
}
    \label{figure:3}
\end{figure*} 
\begin{figure*}[ht]
    \centering
    \includegraphics[scale=0.60, trim=7mm 100mm 10mm 120mm, clip, width=\textwidth]{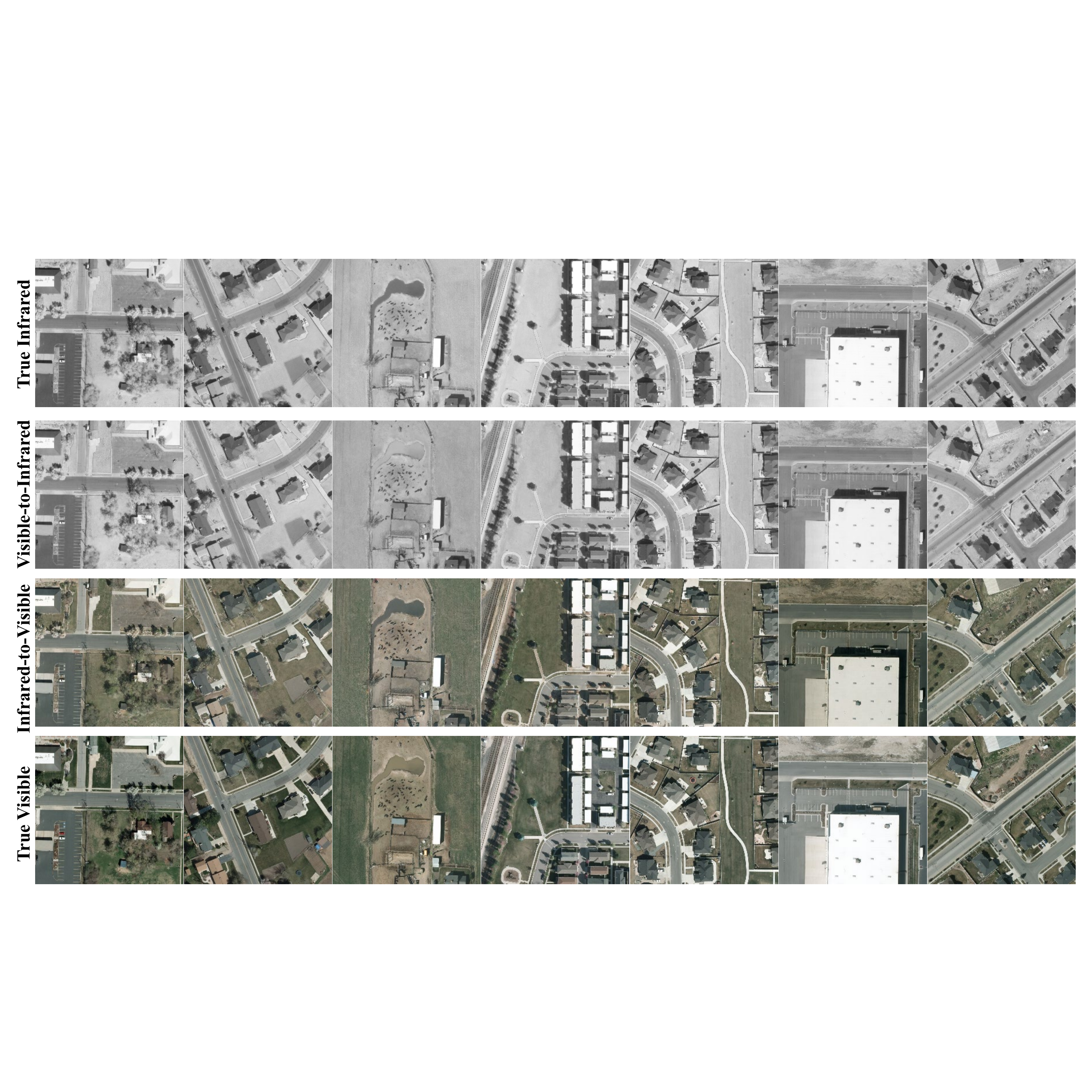}
    \caption{Modality translation results on the VEDAI dataset. The first row represents real infrared images, the second and third rows show generated infrared and visible images, respectively, and the fourth row represents real visible images.
}
    \label{figure:4}
\end{figure*} 

As illustrated in Figs.~\ref{figure:2},~\ref{figure:3}, and~\ref{figure:4}, we provide additional qualitative results on the $\text{M}^{3}\text{FD}$, AVIID, and VEDAI datasets to further validate the effectiveness and generalizability of our proposed method. Each figure showcases representative samples drawn from diverse scenarios.

In these figures, the first row displays the original infrared images, while the fourth row presents the corresponding visible images. The second and third rows illustrate the translated results obtained using the visible-to-infrared and infrared-to-visible translation pathways, respectively. The input conditions and their corresponding outputs are aligned vertically to facilitate clear comparison.

These samples cover a wide range of complex outdoor environments, including urban streets, residential areas, forested regions, wilderness scenes, and water bodies such as lakes and rivers. Despite significant variability in lighting conditions, object scales, textures, and background complexity, our method—CM-Diff—consistently generates visually coherent and semantically aligned translations across both modalities.

The results confirm that CM-Diff exhibits robust cross-modality translation capabilities, maintaining structural consistency and minimizing color distortion even in challenging real-world scenarios. This highlights the model’s strong adaptability and potential for deployment in practical aerial sensing and remote vision applications.
\begin{figure*}[!h]
    \centering
    \includegraphics[scale=0.60, trim=34mm 215mm 39mm 180mm, clip, width=\textwidth]{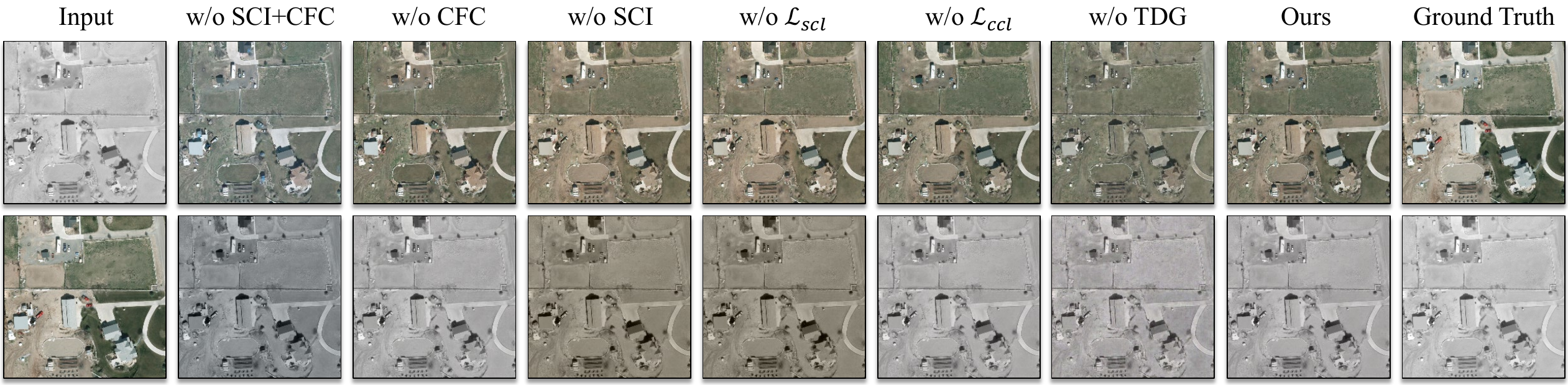}
    \caption{Comparisons of different ablation studies for the visible-to-infrared image translation task. 
    We evaluate the impact of the Statistical Constraint Inference (SCI), which consists of $\mathcal{L}_{{ccl}}$ and $\mathcal{L}_{{scl}}$, the Cross-modality Feature Control (CFC) component, and the translation direction guidance (TDG). }
    \label{figure8}
\end{figure*} 

\subsection{Ablation Visualizations}
To further investigate the contribution of each novel component in our proposed framework, we present qualitative results from a series of ablation experiments conducted on the VEDAI~\cite{VEDAI} dataset. These experiments cover both infrared-to-visible and visible-to-infrared image translation tasks. As illustrated in Fig.~\ref{figure8}, each column corresponds to a different model variant, allowing for direct visual comparison of the impact of specific design choices.

Notably, the translated results shown in the second, fourth, and fifth columns exhibit clear color distortions and unnatural artifacts. These models lack the statistical guidance provided by our proposed constraint mechanism, and consequently, fail to properly regulate the sampling process during generation. This often leads to aberrant color distribution or poor semantic alignment with the target domain. Such visual artifacts are especially pronounced in high-frequency regions, indicating the importance of statistical consistency in maintaining photorealistic quality.

In contrast, the results shown in the seventh column of Fig.~\ref{figure8} correspond to our full model, which integrates both the cross-modality feature interaction mechanism and the statistical constraint inference (SCI) strategy. These images display enhanced color realism, better-preserved object boundaries, and overall superior visual fidelity—closely resembling the ground-truth visible and infrared images. Moreover, the reduced blurriness and clearer contours in these results also validate the effectiveness of incorporating direction label guidance, which facilitates more targeted and stable translation between the two modalities.

Together, these observations underscore the necessity of combining feature-level cross-modality control, distribution-level statistical constraint, and direction-aware supervision to achieve high-quality and consistent image translation in both directions.

\subsection{Implementation Details}
To effectively integrate feature information across infrared and visible modalities, we incorporate a cross-modality attention mechanism at multiple resolutions within both the encoder and decoder pathways of a U-Net backbone. Specifically, attention modules with 64 channels are applied at feature map resolutions of $8 \times 8$, $16 \times 16$, and $32 \times 32$, enabling rich feature interaction across modalities at various semantic levels. To further enhance edge-awareness, we utilize a pre-trained DexiNed model to extract fine-grained edge information from both infrared and visible inputs. These edge maps are fused into the main network to improve structural detail preservation in the final output.

All input images are uniformly resized to a resolution of $256 \times 256$ for both training and evaluation. The U-Net backbone is configured with a base channel width of 128 across all datasets. Training is conducted using the AdamW optimizer with a batch size of 6. The learning rate is initialized to $1 \times 10^{-4}$ and decayed multiplicatively by a factor of 0.9 every 2,000 iterations. The total training procedure spans 100,000 iterations.

To improve the fidelity of the generation process, we modify the original noise schedule by reducing the final noise variance $\beta_T$ from $0.02$ to $0.01$, effectively lowering the amount of noise injected during diffusion and thereby reducing degradation artifacts. The revised noise schedule linearly increases $\beta_t$ from $\beta_1 = 0.0001$ to $\beta_T = 0.01$.

Given the symmetric nature of modality translation between infrared and visible domains, we set both translation loss weights equally as $\lambda_{\textsc{ir}\rightarrow\textsc{vis}} = \lambda_{\textsc{vis}\rightarrow\textsc{ir}} = 1.0$, even though the infrared modality typically presents a lower learning difficulty. During the generative phase, we adopt two constraint-based guidance terms: a content-consistency loss ($\lambda_{\mathrm{ccl}} = 20.0$) and a structure-consistency loss ($\lambda_{\mathrm{scl}} = 20.0$), both of which help to maintain semantic alignment and spatial coherence across modalities.

All experiments are conducted on a computing platform equipped with an Intel(R) Xeon(R) CPU E5-2698 v4 @ 2.20GHz and four NVIDIA V100 GPUs.

\begin{figure*}[t]
    \centering
    \includegraphics[scale=0.54, trim=22mm 0mm 22mm 29mm, clip]{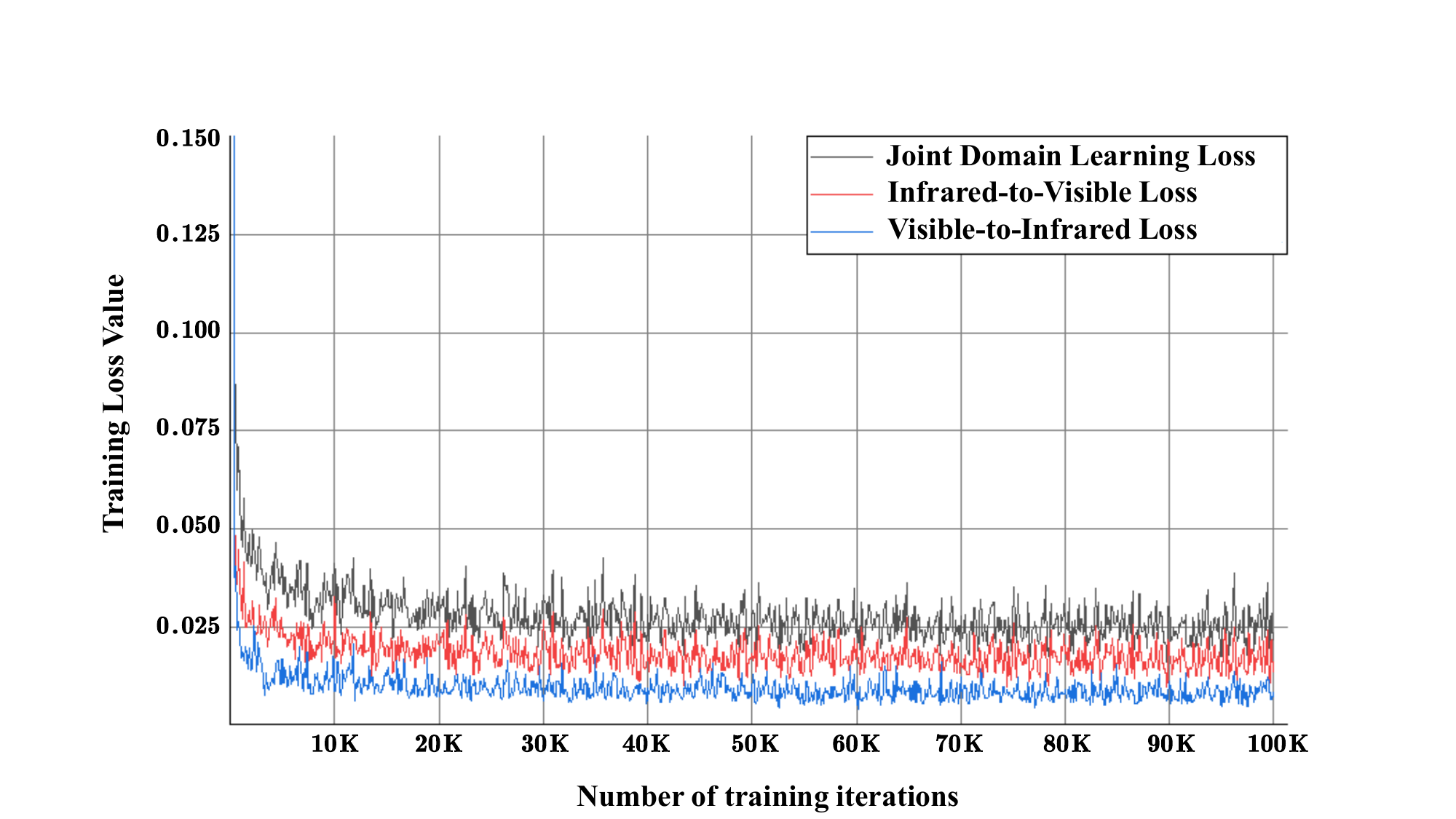}
    \caption{The training loss evolves with the number of iterations. Due to the significant fluctuations in the loss values of individual iterations, which make it challenging to discern overall trends, we instead utilize the average loss value over an entire training epoch and use the iteration count as the horizontal axis.}
    \label{figure6}
\end{figure*} 

\begin{figure*}[t]
    \centering
    \includegraphics[scale=0.9, trim=0mm 132mm 0mm 131mm, clip]{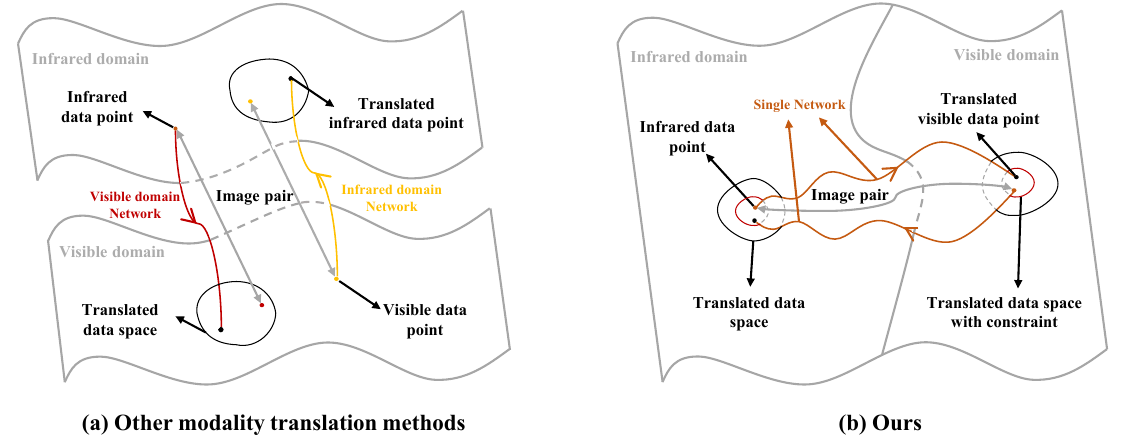}
    \caption{Comparison of bidirectional cross-modality translation frameworks.(a) Other modality translation methods: Independent networks for unidirectional translation (red / yellow arrows) operate in decoupled domains, relying on paired-image constraints but lacking joint optimization. (b) CM-Diff (Ours): A unified network performs simultaneous bidirectional mapping via direction-guided Bidirectional Diffusion Training (BDT) , with Statistical Constraint Inference (SCI) enforcing distributional consistency across modalities ("translated data space with constraint").}
    \label{figure12}
\end{figure*} 

\subsection{Metrics}
To objectively assess the performance of our proposed method on cross-modality image translation tasks, we adopt four widely used image quality assessment metrics: Frechet Inception Distance (FID), Structural Similarity Index Measure (SSIM), Learned Perceptual Image Patch Similarity (LPIPS), and Peak Signal-to-Noise Ratio (PSNR). These metrics provide a comprehensive evaluation from both perceptual and pixel-level perspectives.

\textit{Frechet Inception Distance (FID)}~\cite{FID} quantifies the distance between the feature distributions of real and generated images. Lower FID scores indicate higher similarity between the generated and real image domains in terms of deep features. It is defined as:
\begin{equation}
\text{FID}(v',v) =
\|\mu_v - \mu_{v'}\|^2 + \operatorname{tr}(\Sigma_v + \Sigma_{v'} - 2(\Sigma_v \cdot \Sigma_{v'})^{1/2}),
\end{equation}
where $\mu_v$ and $\mu_{v'}$ are the mean feature vectors of real and generated images, respectively, and $\Sigma_v$ and $\Sigma_{v'}$ are the corresponding covariance matrices. These features are extracted using a pre-trained Inception-V3 network.

\textit{Structural Similarity Index Measure (SSIM)}~\cite{SSIM} evaluates the similarity between the translated and ground-truth images based on luminance, contrast, and structural information. SSIM values range from 0 to 1, with higher scores indicating better perceptual quality. The metric is computed as:
\begin{equation}
\text{SSIM}(v',v) =
\frac{(2\mu_v \cdot \mu_{v'} + c_1)(2\sigma_{v,v'} + c_2)}{(\mu_v^2 + \mu_{v'}^2 + c_1)(\sigma_v^2 + \sigma_{v'}^2 + c_2)},
\end{equation}
where $\mu_v$ and $\mu_{v'}$ denote the mean intensities, $\sigma_v^2$ and $\sigma_{v'}^2$ represent the variances, and $\sigma_{v,v'}$ is the covariance between $v$ and $v'$. $c_1$ and $c_2$ are stability constants.

\textit{Learned Perceptual Image Patch Similarity (LPIPS)}~\cite{LPIPS} measures perceptual similarity by comparing deep feature representations of image patches. It is sensitive to semantic differences and has been shown to correlate well with human perception. LPIPS is computed as:
\begin{equation}
\text{LPIPS}(v',v) =
\sum_{i=1}^{I} \sum_{h_i,w_i} \frac{\omega_i \cdot \left[f_i(v')_{h_i,w_i} - f_i(v)_{h_i,w_i}\right]^2}{\mathcal{H}_i \cdot \mathcal{W}_i},
\end{equation}
where $f_i(\cdot)$ denotes the features extracted from the $i$-th layer of a pre-trained AlexNet~\cite{krizhevsky2012imagenet}, and $\omega_i$ is the learned weight for the $i$-th layer. $\mathcal{H}_i$ and $\mathcal{W}_i$ denote the spatial dimensions of the $i$-th layer's feature map.

\textit{Peak Signal-to-Noise Ratio (PSNR)} evaluates image fidelity at the pixel level by measuring the logarithmic ratio between the maximum possible pixel value and the mean squared error (MSE) between the real and translated images. It is defined as:
\begin{equation}
\text{PSNR}(v',v) =
10 \cdot \log_{10} \left( \frac{\mathcal{M} \cdot \mathcal{N} \cdot \text{INTENSITY}_{\text{max}}^2}{\sum_{i=1}^{\mathcal{M}} \sum_{j=1}^{\mathcal{N}} [v_{ij} - v'_{ij}]^2} \right),
\end{equation}
where $\mathcal{M}$ and $\mathcal{N}$ are the image height and width, respectively, and $\text{INTENSITY}_{\text{max}}$ is the maximum pixel value, which is 255 for 8-bit images.

\subsection{Datasets} 
\textit{The VEDAI dataset}~\cite{VEDAI} is designed for dual-modality small vehicle detection in aerial imagery. It contains a total of 1,209 paired infrared-visible image samples with resolutions of $1024 \times 1024$ and $512 \times 512$, captured under varied scene conditions. The infrared modality is based on near-infrared (NIR) imaging, making the dataset particularly suitable for cross-modality image translation and detection tasks involving small objects with complex backgrounds. The dataset also provides comprehensive pixel-level annotations, which facilitate both supervised training and objective performance benchmarking. In our experiments, we allocate 1,089 image pairs for training and the remaining 120 for testing to ensure a fair and balanced evaluation.

\textit{The AVIID dataset}~\cite{AVIID} consists of paired visible-infrared images captured by a dual-spectrum camera mounted on an unmanned aerial vehicle (UAV), across varying illumination and environmental conditions. It is subdivided into three subsets: AVIID-1 includes 993 image pairs of resolution $434 \times 434$ featuring road scenes and common vehicles; AVIID-2 contains 1,090 low-light image pairs with visible noise and motion blur; and AVIID-3 offers 1,280 image pairs at $512 \times 512$ resolution, covering diverse scenarios such as roads, bridges, and residential streets. While the dataset was originally constructed for the visible-to-infrared image translation task, our framework supports bidirectional translation. Therefore, we adopt the full dataset for both IR$\rightarrow$VIS and VIS$\rightarrow$IR tasks. We divide the dataset into 2,674 training pairs and 669 testing pairs.

\textit{The $\text{M}^{3}\text{FD}$ dataset}~\cite{M3FDdataset} provides 4,500 spatially registered visible and infrared image pairs with a resolution of $1024 \times 768$. Although initially designed for multi-sensor fusion and object detection tasks, the high-resolution and scene diversity of the dataset make it well-suited for image-to-image translation applications. The data is collected from multiple urban and suburban locations in Dalian, China, including the Dalian University of Technology, Golden Stone Beach, and Jinzhou District. Despite slight misalignments between the modalities, the dataset remains valuable for training robust translation models. For our experiments, we utilize 3,780 image pairs for training and reserve 210 pairs for testing.

\begin{table}[ht]     \caption{Code and Dataset Sources for Evaluated Methods}
    \centering
    \resizebox{0.48\textwidth}{!}{
    \begin{tabular}{c|c}
        \toprule
         Method &  Code address \\
         \midrule
         T2V-DDPM & \url{https://github.com/Nithin-GK/T2V-DDPM} \\ 
         CycleGAN & \url{https://github.com/junyanz/pytorch-CycleGAN-and-pix2pix} \\ 
         UNIT & \url{https://github.com/mingyuliutw/UNIT} \\ 
         MUNIT & \url{https://github.com/NVlabs/MUNIT} \\ 
         DCLGAN & \url{https://github.com/JunlinHan/DCLGAN} \\ 
         I2V-GAN & \url{https://github.com/BIT-DA/I2V-GAN} \\
         ROMA & \url{https://github.com/BIT-DA/ROMA} \\
         CUT & \url{https://github.com/taesungp/contrastive-unpaired-translation} \\
        \midrule
         Dataset & Data address \\
         \midrule
         AVIID & \url{Averial-visible-to-infrared-image-translation} \\
         VEDAI & \url{https://downloads.greyc.fr/vedai/} \\
         $\text{M}^{3}\text{FD}$ & \url{https://github.com/dlut-dimt/TarDAL} \\
    \bottomrule
    \end{tabular}
    }
    \label{tab:5}
\end{table}

\bibliography{aaai2026}


\end{document}